\documentclass[11pt,letterpaper]{article}

\usepackage[letterpaper,margin=0.98in]{geometry}
\usepackage[utf8]{inputenc}
\usepackage[T1]{fontenc}
\usepackage{microtype}
\usepackage{url}

\usepackage{amsmath}
\usepackage{amssymb}
\usepackage{amsfonts}
\usepackage{amsthm}
\usepackage{mathtools}
\usepackage{nicefrac}

\usepackage{graphicx}
\usepackage{subcaption}
\usepackage{wrapfig}
\usepackage{booktabs}
\usepackage{multirow}

\usepackage[table,xcdraw]{xcolor}
\definecolor{darkgray}{RGB}{50,50,60}
\definecolor{rulegray}{RGB}{180,190,220}
\definecolor{highlight}{gray}{0.92}
\definecolor{lightblue}{RGB}{100,149,237}
\definecolor{sienna}{RGB}{136,45,23}
\definecolor{slategray}{RGB}{90,90,90}
\definecolor{burgundy}{RGB}{128,0,32}
\definecolor{medblue}{RGB}{70,130,180}
\definecolor{ourcolor}{RGB}{235,242,255}
\definecolor{untempcolor}{RGB}{255,240,230}

\definecolor{thmbg}{RGB}{243,245,248}        
\definecolor{thmrule}{RGB}{52,82,115}        
\definecolor{lemmabg}{RGB}{243,245,248}      
\definecolor{lemmarule}{RGB}{75,95,120}      
\definecolor{assbg}{RGB}{243,245,248}        
\definecolor{assrule}{RGB}{105,120,140}      
\definecolor{takeawaybg}{RGB}{243,245,248}   
\definecolor{takeawayrule}{RGB}{90,90,95}    

\usepackage[normalem]{ulem}
\useunder{\uline}{\ul}{}
\usepackage{enumitem}

\usepackage{authblk}

\usepackage{tcolorbox}
\tcbuselibrary{skins, breakable}
\usepackage{mdframed}

\newtcolorbox{thmboxinner}{
  enhanced, breakable, rounded corners,
  colback=thmbg, colframe=thmrule,
  boxrule=0pt, leftrule=2.5pt,
  left=10pt, right=8pt, top=4pt, bottom=4pt,
  before skip=8pt, after skip=8pt,
}
\newtcolorbox{lemmaboxinner}{
  enhanced, breakable, rounded corners,
  colback=lemmabg, colframe=lemmarule,
  boxrule=0pt, leftrule=2.5pt,
  left=10pt, right=8pt, top=4pt, bottom=4pt,
  before skip=8pt, after skip=8pt,
}
\newtcolorbox{assumptionboxinner}{
  enhanced, breakable, rounded corners,
  colback=assbg, colframe=assrule,
  boxrule=0pt, leftrule=2.5pt,
  left=10pt, right=8pt, top=4pt, bottom=4pt,
  before skip=8pt, after skip=8pt,
}

\newtcolorbox{takeawaybox}{
  enhanced, breakable, rounded corners,
  colback=takeawaybg, colframe=takeawayrule,
  boxrule=0pt, leftrule=2.5pt,
  left=10pt, right=8pt, top=4pt, bottom=4pt,
  before skip=8pt, after skip=8pt,
  fontupper=\small\itshape,
}

\usepackage{algorithm}
\usepackage{algorithmic}

\usepackage[textsize=tiny]{todonotes}

\newtheorem{theorem}{Theorem}
\theoremstyle{plain}
\newtheorem{lemma}{Lemma}

\theoremstyle{definition}

\newtheorem{assumption}{Assumption}

\newenvironment{assumptionbox}
  {\begin{assumptionboxinner}\begin{assumption}}
  {\end{assumption}\end{assumptionboxinner}}

\usepackage{amsmath,amsfonts,bm}

\def\eqref#1{equation~\ref{#1}}

\def\1{\bm{1}}

\def\ve{{\bm{e}}}

\def\vp{{\bm{p}}}
\def\vq{{\bm{q}}}

\def\mE{{\bm{E}}}

\def\mI{{\bm{I}}}

\DeclareMathAlphabet{\mathsfit}{\encodingdefault}{\sfdefault}{m}{sl}
\SetMathAlphabet{\mathsfit}{bold}{\encodingdefault}{\sfdefault}{bx}{n}

\def\gD{{\mathcal{D}}}
\def\gE{{\mathcal{E}}}

\def\gM{{\mathcal{M}}}

\def\gU{{\mathcal{U}}}
\def\gV{{\mathcal{V}}}

\newcommand{\err}[1]{{\tiny$\pm$#1}}
\newcommand{\appref}[1]{Appendix~\ref{#1}}

\newcommand{\takeaway}[1]{%
  \begin{takeawaybox}#1\end{takeawaybox}}

\usepackage[
  style=authoryear-comp,
  maxbibnames=15,
  maxcitenames=2,
  giveninits=true,
  natbib=true,
  backend=biber,
  sorting=nyt,
  uniquename=false,
  uniquelist=false,
]{biblatex}
\usepackage{hyperref}
\hypersetup{
  colorlinks=true,
  breaklinks=true,
  citecolor=medblue,
  linkcolor=sienna,
  urlcolor=sienna,
  filecolor=black,
  bookmarksopen=false,
}

\DeclareFieldFormat{citehyperref}{%
  \DeclareFieldAlias{bibhyperref}{noformat}%
  \bibhyperref{#1}}
\DeclareFieldFormat{textcitehyperref}{%
  \DeclareFieldAlias{bibhyperref}{noformat}%
  \bibhyperref{%
    #1%
    \ifbool{cbx:parens}
      {\bibcloseparen\global\boolfalse{cbx:parens}}
      {}}}

\savebibmacro{cite}
\savebibmacro{textcite}

\renewbibmacro*{cite}{%
  \printtext[citehyperref]{%
    \restorebibmacro{cite}%
    \usebibmacro{cite}}}

\renewbibmacro*{textcite}{%
  \ifboolexpr{
    ( not test {\iffieldundef{prenote}}
      and test {\ifnumequal{\value{citecount}}{1}} )
    or
    ( not test {\iffieldundef{postnote}}
      and test {\ifnumequal{\value{citecount}}{\value{citetotal}}} )
  }
    {\DeclareFieldAlias{textcitehyperref}{noformat}}
    {}%
  \printtext[textcitehyperref]{%
    \restorebibmacro{textcite}%
    \usebibmacro{textcite}}}
\usepackage[capitalize,noabbrev]{cleveref}
\renewcommand{\citet}[1]{\citeauthor{#1}~(\citeyear{#1})}

\crefname{theorem}{Theorem}{Theorems}        \Crefname{theorem}{Theorem}{Theorems}
\crefname{lemma}{Lemma}{Lemmas}              \Crefname{lemma}{Lemma}{Lemmas}
\crefname{proposition}{Proposition}{Propositions}
\Crefname{proposition}{Proposition}{Propositions}
\crefname{corollary}{Corollary}{Corollaries} \Crefname{corollary}{Corollary}{Corollaries}
\crefname{definition}{Definition}{Definitions}
\Crefname{definition}{Definition}{Definitions}
\crefname{assumption}{Assumption}{Assumptions}
\Crefname{assumption}{Assumption}{Assumptions}
\crefname{remark}{Remark}{Remarks}           \Crefname{remark}{Remark}{Remarks}
\crefname{example}{Example}{Examples}        \Crefname{example}{Example}{Examples}
\crefname{algorithm}{Algorithm}{Algorithms}  \Crefname{algorithm}{Algorithm}{Algorithms}

\title{Entropy Aware Reward Guidance for
  Diffusion \\ Language Model Alignment \vspace{3mm}
  \footnotetext{Corresponding author: \texttt{atutej@utexas.edu}}}
\author{Atula Tejaswi\textsuperscript{$\ast$}}
\author{Litu Rout\textsuperscript{$\ast$}}
\author{Constantine Caramanis}
\author{Sanjay Shakkottai}
\author{Sujay Sanghavi}
\affil{The University of Texas at Austin}
\date{}

\begin{document}

\maketitle

\begin{abstract}
\noindent Reward guidance, also known as posterior sampling, is a popular method for test-time adaptation and post-training in continuous diffusion models. In this paper, we study reward guidance for \emph{discrete} diffusion language models; now, one cannot differentiate through the natural outputs of the model because they are discrete tokens. We introduce a novel mechanism called \textit{EntRGi: \underline{Ent}ropy aware \underline{R}eward \underline{G}u\underline{i}dance} to address this issue. EntRGi dynamically interpolates between continuous token relaxations and sampled hard tokens, on a token-by-token basis, using the diffusion model's predictive entropy. We demonstrate that EntRGi maintains both reward model reliability and optimization accuracy, while existing approaches sacrifice one for the other. We empirically validate our approach on 7B-parameter diffusion language models across two settings: (1) test-time adaptation, and (2) \textit{RGRL: \underline{R}eward \underline{G}uided \underline{R}einforcement \underline{L}earning} , our recipe for post-training on reward-guided data, showing consistent improvements over state-of-the-art methods. Our code is available at \url{https://atutej.github.io/entrgi-rgrl}.
\end{abstract}

\section{Introduction}
\label{sec:intro}

Reward guidance has proven highly effective for adapting continuous diffusion models, where feedback from a downstream reward model is used to iteratively refine each denoising step toward desired outcomes~\citep{cg, prabhudesai2024videodiffusionalignmentreward, freedom, mpgd, tfg}. This paradigm has enabled controllable generation across inverse problems~\citep{dps, p2l, psld, stsl}, stylization~\citep{stylealigned, rbm}, and semantic editing~\citep{rfi}, allowing diffusion models to optimize task-specific objectives without retraining.

In this work, we study reward guidance in the setting of \emph{discrete} diffusion large language models (dLLMs)~\citep{d3pm, sedd, mdlm, md4, llada, dream, geminidiffusion}. Unlike autoregressive LLMs, dLLMs generate text by starting from a fully masked sequence and iteratively denoising tokens in parallel, not necessarily committing to a fixed left-to-right order. Iterative denoising allows for a naive method for reward-guided adaptation based on particle filtering style approaches: at any step, generate many different noisy completions, and then reject the ones with low reward \citep{pg-dlm, fk-steering, ou2025inference}. 

Our work is motivated by a common setting in continuous diffusion, where downstream reward models are {\em differentiable} in their inputs (which are the outputs of the diffusion model)  ~\citep{cg}. In that setting, gradient feedback from the downstream rewards results in a strong signal that is used to change the iterative refinement process with much higher efficacy. {\bf Our motivation} is to realize a similar powerful effect for diffusion language models, where the outputs are discrete tokens and reward models are themselves fine-tuned language models. 

The discrete nature of dLLM outputs presents a natural challenge: it prevents direct gradient propagation as is done in continuous settings. A natural fix is to replace discrete tokens with continuous embeddings to enable gradient flow; however reward models have not seen such soft out-of-vocabulary inputs in their training \citep{g2d2, tess-2, wang2025finetuningdiscretediffusionmodels}, hurting performance. A recent method, APS \citep{aps} proposes a framework to mitigate this issue. APS evaluates the reward at sampled hard tokens (from the reward model vocabulary), but propagates gradients as if the input were soft via the straight-through estimator (STE)~\citep{ste, gumbel-soft}. This second approach however introduces a mismatch between where the reward is evaluated and where gradients are applied. 
Thus, both of these existing approaches  tradeoff between gradient accuracy and reward model reliability. Resolving this tension is essential not only for stronger inference-time steering, but also for unlocking dense reward-gradient feedback as a viable post-training alternative to supervised fine-tuning \citep{llada} and scalar-reward RL methods~\citep{d1}.

To address this tradeoff, we introduce \textbf{EntRGi} (\textbf{Ent}ropy-aware \textbf{R}eward \textbf{G}u\textbf{i}dance), an entropy-aware reward guidance mechanism for discrete diffusion language models. EntRGi explores the following question: How can we {\em effectively} leverage reward gradients to iteratively guide a discrete diffusion LLM generation toward higher-reward token sequences? As illustrated in \autoref{fig:entrgi_main}, EntRGi adaptively interpolates between continuous token embeddings and sampled hard token embeddings using the dLLM's own per-token entropy: soft inputs are favored when the model is confident, and hard inputs when it is uncertain. This simple mechanism provides reliable gradients during optimization while ensuring the reward model is evaluated on inputs it can interpret throughout the denoising process. We further show that these reward-guided samples can be used in a novel post-training algorithm \textbf{RGRL}: \textbf{R}eward \textbf{G}uided \textbf{R}einforcement \textbf{L}earning.

{\bf Our contributions} can be summarized as follows:
\textbf{(1)} We introduce EntRGi, an entropy-aware reward guidance mechanism for discrete dLLMs.
\textbf{(2)} In the test-time adaptation setting, we demonstrate that EntRGi outperforms APS \citep{aps}, the prior state-of-the-art.
\textbf{(3)} We develop a new post-training recipe: RGRL that generates reward-gradient guided samples and fine-tunes on those. This is as opposed to standard RL methods that do not guide generation. We show our method significantly outperforms (70\% relative improvement) diffu-GRPO \citep{d1}, a widely adopted RL algorithm for dLLMs. This recipe works with both APS guidance and EntRGi guidance, with the latter providing higher gains.

We demonstrate these contributions with experiments on up to two 7B+ parameter models \citep{dream, llada}, 5 multi-skill datasets \citep{rewardbench2, judgebench, rmbench, zhao2024wildchat1mchatgptinteraction}, and 4 reward models~\citep{skywork}, analyzing the mechanisms underlying the improvements over prior methods.

\section{Related Work}
\label{sec:rel-work}

\textbf{Discrete diffusion posterior sampling.}
Discrete diffusion models offer a non-autoregressive alternative for posterior sampling over categorical sequences, generating predictive distributions over all tokens in parallel at each denoising step. This makes them well-suited for posterior sampling under external constraints such as reward models, without retraining or task-specific fine-tuning~\citep{pg-dlm, aps}.

\textbf{Reward-gradient-free methods.}
These methods avoid back-propagating through the reward model and rely only on scalar reward queries. At inference time, search-based and particle methods have been extensively developed for continuous diffusion~\citep{dts, dm_noise_trajectory, guo2025trainingfreeguidancedifferentiabilityscalable, kim2025testtimealignmentdiffusionmodels, zhang2025inferencetimescalingdiffusionmodels}, with recent extensions to discrete diffusion including Best-of-$N$ and particle-based sampling~\citep{pg-dlm, sgdd, fk-steering, ou2025inference, svdd}. At training time, RL-based fine-tuning treats denoising as an MDP, and applies policy gradient methods using scalar, non-differentiable rewards, an approach well-developed for continuous diffusion~\citep{black2024trainingdiffusionmodelsreinforcement, fan2023dpokreinforcementlearningfinetuning} and recently extended to dLLMs~\citep{d1, zekri2025finetuningdiscretediffusionmodels}. Posterior matching~\citep{ddpp} and preference-based methods~\citep{d2po, tr2d2} sidestep reward gradients through preference-based training objectives. While avoiding gradient approximation, these methods often suffer from sample inefficiency or slow convergence~\citep{aps, g2d2}.

\textbf{Reward-gradient-based methods.}
These methods back-propagate gradients through a differentiable reward model. Gradient-based reward guidance is extensively studied in continuous diffusion, both for inference-time steering~\citep{cg, dps, bansal2023universalguidancediffusionmodels} and for fine-tuning~\citep{clark2024directlyfinetuningdiffusionmodels, prabhudesai2024videodiffusionalignmentreward, xu2023imagerewardlearningevaluatinghuman, anil2026finetuningdiffusionmodelsintermediate}, but remains comparatively less developed for discrete diffusion. Existing methods either feed continuous relaxations of token embeddings to the reward model~\citep{g2d2, tess-2}, querying the reward out-of-distribution, or discretize via the straight-through estimator (STE)~\citep{ste, gumbel-soft}, propagating gradients evaluated at sampled hard tokens: APS~\citep{aps} is the prior state-of-the-art for inference-time steering of dLLMs, and DRAKES~\citep{wang2025finetuningdiscretediffusionmodels} adopts the straight-through mechanism for fine-tuning of biological sequence diffusion models.

\textbf{Challenges and limitations.}
Both relaxation and STE-based methods face a fundamental challenge: the mismatch between discrete model outputs and the continuous representations required for gradient propagation, which is most pronounced during early denoising steps when per-token predictive entropy is high. To the best of our knowledge, ours is the first work in dLLMs to leverage model uncertainty for adaptive gradient regulation at inference time. At training time, we further use these reward-guided completions to provide dense feedback, going beyond scalar-reward RL.

\section{Reward Guidance for Discrete Diffusion LLMs}
\label{sec:method}

\textbf{Preliminaries.} Discrete diffusion language models \citep{mdlm, sedd, llada, dream} are generative models that operate over $L$-length sequences of tokens drawn from a finite vocabulary $\gV$. A commonly used instantiation is the masked diffusion setting, where each token is from a vocabulary consisting of $K$ ``actual'' tokens and one ``mask'' token $m$. Standard generation (i.e. the ``reverse process'') in masked diffusion starts from time $T$ and an initial string of all masks $z_T = m^L$. Time goes from $t=T$ to $t=0$, and each $z_{t-1}$ is made from the preceding $z_t$ by first choosing $k$ currently masked tokens in $z_t$ and unmasking them using the probability distribution from one inference pass of the diffusion model. It ends with a string $z_0$ that contains no mask tokens.
We now develop notations to make this specific. 

Let $\gM_t$ be the set of masked positions in $z_t$. In this work we focus on the ``unmask and commit'' mode of generation \citep{mdlm}, which means that that once a token is unmasked it remains fixed for all subsequent steps. That means that $z_{t-1}^l=z_t^l \quad \text{for all $l\notin \gM_t$}$.

For the currently masked positions, we input $z_t$ into the diffusion model to obtain logits that we will sample from. Let $\theta$ denote the parameters of the diffusion model. For any currently masked position $l\in \gM_t$, define $\vp_\theta^l(z_t)$  to be the resulting probability distribution over the vocabulary. Finally, let  $\vq^{\gM_t} = \vp^{\gM_t}_\theta(z_t)$ denote the set of distributions over all currently masked locations $l \in \gM_t$. 

The first step in unmasking is to choose a set $\gU(\vq^{\gM_t})$ of currently-masked tokens according to some pre-set selection logic. For example, in the models \textit{Dream-v0-Instruct-7B}~\citep{dream} and \textit{LLaDA-8B-Instruct}~\citep{llada} used in this work, this pre-set selection logic is to pick a few tokens whose distributions $\vq^l$ have the smallest entropy. Once we have this set $\gU(\vq^{\gM_t})$, we generate the remaining tokens in $z_{t-1}$ by sampling tokens in $\mathcal{U}(\vq^{\gM_t})$ i.e.
$z_{t-1}^l ~ \sim ~ \vq^l \quad \text{for $l\in \mathcal{U}(\vq^{\gM_t})$}$
and keeping all the other tokens as masks, i.e. $z_{t-1}^l = m$ for all $l \in \gM_t \setminus \gU(\vq^{\gM_t})$.

\subsection{Entropy Aware Reward Guidance}
Recall that we want to change the above generation process so that it is more likely to generate high reward strings as measured by a downstream reward model $R$. Typically, $R$ is itself a language model fine-tuned to output scalar scores \citep{skywork, armorm, rlhf_openai}. For now, let us assume that the vocabulary of the reward model consists of the same $K$ ``actual'' tokens as that of the diffusion model vocabulary $\mathcal{V}$ (we relax this later). Naively, the input to $R$ is a string of $L$ discrete tokens $x = (x^1,\ldots, x^l, \ldots, x^L)$. Note that during inference in $R$, every token $x^l$ is immediately converted into an embedding vector $\mE^R[x^l]$ i.e. by looking up each token in the input embedding table $\mE^R$ of the model $R$.

In this work we will find it useful to treat $R$ more generally as a scalar function of $L$ input {\em embedding vectors} $\ve^1,\ldots,\ve^L$, each of which {\em may or may not} be members of the input embedding table $\mE^R$. We denote this (more general) function as $R(\ve)$ where $\ve = (\ve^1,\ldots,\ve^L)$. We assume that $R(\ve)$ is a differentiable function of the vectors $\ve$.

As shown in \autoref{fig:entrgi_main}, at each masked position $l \in \gM_t$, the diffusion model produces a predicted distribution $\vq^l$. From $\vq^l$, we can obtain a hard token
${\ve}_\text{hard}^l = \mE^R[x^l]$, $x^l \sim \vq^l$.
Then, we want to update $\vq^l$ to increase $R$.
To do so, we would like feed $R$ the sequence of embeddings ${\ve}_\text{hard}$ and update $\vq^l$ in a direction that improves reward. However, sampling breaks differentiability, so we will instead feed the reward model a differentiable surrogate input that preserves gradient flow to $\vq^l$. Prior works explore the following choices of inputs to the reward:

\begin{figure*}[t]
  \centering
  \includegraphics[width=\linewidth]{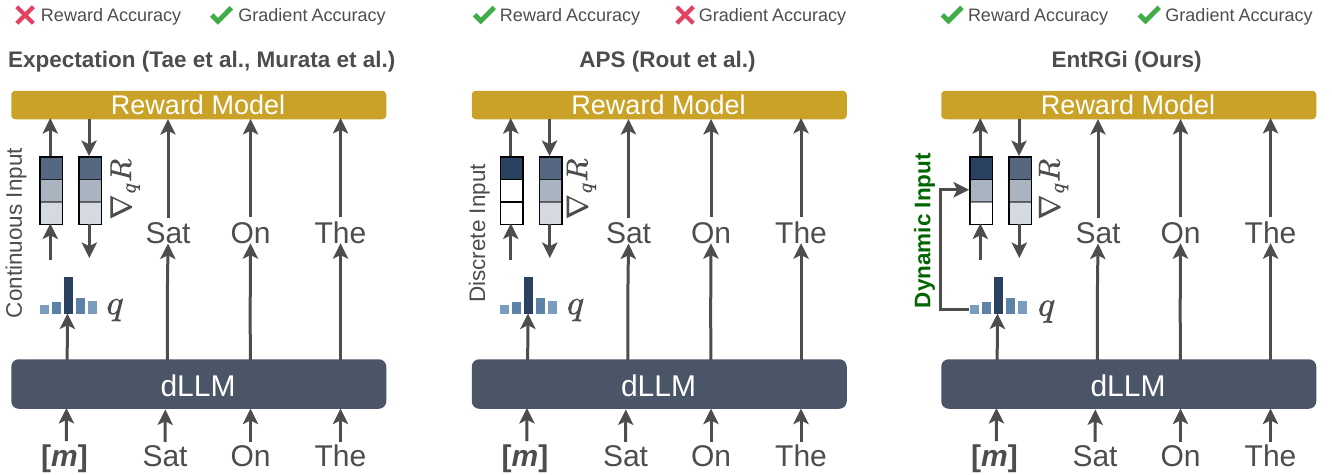}
  \caption{\textbf{Overview of \textbf{Ent}ropy-aware \textbf{R}eward \textbf{G}u\textbf{i}dance (EntRGi)}. Given logits $\vq$ from a diffusion language model (dLLM) at masked positions [$m$], the goal is to update them to maximize a reward $R$ defined by a reward model. Prior work either compromises reward reliability \citep{tess-2, g2d2}, or gradient accuracy \citep{aps}. EntRGi addresses these limitations by constructing inputs as a dynamic \emph{entropy-weighted interpolation} between continuous token embeddings and sampled hard tokens.}
  \label{fig:entrgi_main}
\end{figure*}

\textbf{Feeding the Expectation.} These approaches~\citep{tess-2, g2d2} feed $R$ the ``expected'' soft embedding ${\ve}_\text{soft}^l = \sum_{k=1}^K \vq^l_k\,\mE^R_k$, assuming it can reliably interpret such inputs. This yields $\nabla_{\vq^l} R({\ve}_\text{soft}) = (\partial R/\partial{\ve}_\text{soft}^l) \cdot (\mE^R)^\top$, with gradients flowing cleanly through ${\ve}_\text{soft}^l$. Here, $\partial R/\partial {\ve}_\text{soft}^l \in \mathbb{R}^{1 \times d}$ is the gradient of the reward with respect to ${\ve}_\text{soft}^l$, $\mE^R \in \mathbb{R}^{K \times d}$ is the embedding matrix. However, since $R$ is trained only on hard-token embeddings, its feedback is reliable only when ${\ve}_\text{soft}^l$ lies close to a real token embedding. We therefore define the \textbf{vocabulary error} as $\mathcal{D}^l = \min_k\|{\ve}_\text{soft}^l - \mE^R_k\|$. As $H(\vq^l)$ increases, ${\ve}_\text{soft}^l$ drifts from any single token, so $\mathcal{D}^l$ grows and the gradient becomes unreliable.

\textbf{Feeding a hard token embedding to $R$.} To resolve this out-of-distribution mismatch, APS \citep{aps} fixes $\mathcal{D}^l = 0$ by computing inputs $\ve_\text{APS}$ as the following,
\begin{equation}    
\ve_\text{APS}^l =
{\ve}_\text{soft}^l + \mathrm{sg}({\ve}_\text{hard}^l - {\ve}_\text{soft}^l) = \ve^l_\text{hard}
\label{eq:APS}
\end{equation}

where $\mathrm{sg}$ \citep{ste, gumbel-soft} is the stop-gradient operator. The reward is thus evaluated at ${\ve}_\text{hard}^l$ but routed through ${\ve}_\text{soft}^l$, giving $\nabla_{\vq^l} R({\ve}_\text{APS}) = (\partial R/\partial{\ve}_\text{APS}^l)|_{{\ve}_\text{hard}^l} \cdot (\partial{\ve}_\text{APS}/\partial{\ve}_\text{soft}) \cdot (\mE^R)^\top = (\partial R/\partial{\ve}_\text{APS}^l)|_{{\ve}_\text{hard}^l} \cdot (\mE^R)^\top$, since $\partial{\ve}_\text{APS}/\partial{\ve}_\text{soft}=\mI$ from~\autoref{eq:APS}.

The mismatch between reward evaluation at ${\ve}_\text{hard}$ and gradient propagation via ${\ve}_\text{soft}$ induces what we define as the \textbf{approximation error}
$\mathcal{E}^l = \|{\ve}_\text{hard}^l - {\ve}_\text{soft}^l\|$.
Since ${\ve}_\text{hard}^l \sim \vq^l$ and ${\ve}_\text{soft}^l = \mathbb{E}_{\vq^l}[{\ve}_\text{hard}^l]$,
$\mathcal{E}^l = 0$ iff $\vq^l$ is a point mass, and grows as $\vq^l$ spreads, corrupting the gradient at high entropy.

\textbf{EntRGi.} Both failure modes stem from a fundamental tension: $\mathcal{D}^l$ demands ${\ve}_\text{APS}^l$
close to a real token, while $\mathcal{E}^l$ demands ${\ve}_\text{hard}^l$ close to ${\ve}_\text{soft}^l$.
We resolve this by introducing an \emph{adaptive} interpolation weight $w^l \in [0,1]$
and constructing the reward model input $\ve_\text{EntRGi}$ as a convex combination:
\begin{equation}
  {\ve}_\text{EntRGi}^l = {\ve}_\text{soft}^l + \mathrm{sg}\!\bigl(w^l({\ve}_\text{hard}^l - {\ve}_\text{soft}^l)\bigr)
  = (1-w^l){\ve}_\text{soft}^l + w^l{\ve}_\text{hard}^l,
  \label{eq:ehat}
\end{equation}
This unifies the two prior approaches: $w^l=0$ recovers the Expectation method and $w^l=1$ recovers APS, with EntRGi adaptively choosing $w^l$ in between. Then, expanding from \autoref{eq:ehat}:
\begin{align}
  \mathcal{E}^l &= \|{\ve}_\text{EntRGi}^l - {\ve}_\text{soft}^l\| = w^l \cdot \mathcal{E}^l_\text{APS},
  \label{eq:E}\\[-2pt]
  \mathcal{D}^l &= \min_k\|{\ve}_\text{EntRGi}^l - \mE^R_k\|
    \leq \|{\ve}_\text{EntRGi}^l - {\ve}_\text{hard}^l\| = (1-w^l)\|{\ve}_\text{hard}^l - {\ve}_\text{soft}^l\|.
  \label{eq:D}
\end{align}
\begin{assumptionbox}
The deviation $|R({\ve}_\text{EntRGi}^l) - R(\mE^R_{k^*})|$ is small when $\mathcal{D}^l$ is small and grows monotonically with $\mathcal{D}^l$, where $k^* = \arg\min_k \|{\ve}_\text{EntRGi}^l - \mE^R_k\|$.
\label{ass:deviation}
\end{assumptionbox}

\textit{Intuition of~\Cref{ass:deviation}} Since $R$ is trained only on hard-token embeddings, its outputs are reliable near real tokens and degrade as inputs drift away. Empirically, this is consistent with Expectation methods, where feeding $R$ inputs farther from real tokens still yields useful but lower performance.

From \autoref{eq:E}, any $w^l < 1$ gives $\mathcal{E}^l < \mathcal{E}^l_\text{APS}$.
From \autoref{eq:D} and Assumption~1, $(1-w^l)\|{\ve}_\text{hard}^l - {\ve}_\text{soft}^l\|$ must be small.
Since $\|{\ve}_\text{hard}^l - {\ve}_\text{soft}^l\|$ grows with $H(\vq^l)$, this cost is
small at low entropy for any $w^l$ but grows at high entropy, so $w^l$ must grow with
$H(\vq^l)$. Setting $w^l = H(\vq^l)/\log K$ gives\footnote{We ablate alternative weighting mechanisms in \appref{sec:appdx_weighting}.},
\[
  \mathcal{E}^l_\text{EntRGi} = \tfrac{H(\vq^l)}{\log K}\cdot\mathcal{E}^l_\text{APS}
  < \mathcal{E}^l_\text{APS}, \qquad
  \mathcal{D}^l_\text{EntRGi} \leq \bigl(1 - \tfrac{H(\vq^l)}{\log K}\bigr)
  \|{\ve}_\text{hard}^l - {\ve}_\text{soft}^l\|.
\]
Therefore, EntRGi strictly reduces $\mathcal{E}^l$ below APS at every entropy level. $\mathcal{D}^l$ is bounded by $(1-w^l)\cdot\mathcal{E}^l_\text{APS}$,
which is small at low entropy. At high entropy, $w^l \to 1$ so $\mathcal{D}^l \to 0$ by construction. By Assumption~1, keeping $\mathcal{D}^l$ small maintains gradient reliability, so the reduction in $\mathcal{E}^l$ is a strict improvement over APS at every entropy level.

\begin{algorithm}[h]
\caption{EntRGi: Entropy Aware Reward Guidance}
\label{alg:EntRGi}
\begin{algorithmic}[1]
\REQUIRE Reward model $R$, guidance scale $\eta$, reward model gradient steps $M$
\STATE Initialize  blank canvas $z_T = m^L$, and set of masked positions $\gM_T = [L]$  
\FOR{time steps $t = T, T-1, \dots, 1$}  
    \STATE Compute dLLM output distributions $\vq^{\gM_t} = \{ \vq^l, l\in \gM_t \}$
    \FOR{$j = 1, \dots, M$}
         \STATE Compute the soft expected embeddings $\ve^{\gM_t}_\text{soft}$ for masked positions $\gM_t$ using $\vq^{\gM_t}$
        \STATE Sample $x^l \sim \vq^l$ for each $l \in \mathcal{M}_t$, and let the embeddings of this sampled seq be $\ve_\text{hard}^{\gM_t}$
        \STATE Compute per-token weights  $w^l = H(\vq^l) / \log K$ for $l \in \gM_t$  
        \STATE Construct the input to the reward model; \\
        $\ve_\text{EntRGi}^l = \begin{cases}
        \ve_\text{soft}^l + \mathrm{sg}\bigl(w^l (\ve_\text{hard}^l - \ve_\text{soft}^l)\bigr) & l \in \gM_t \\[4pt]
        \ve_\text{hard}^l = \text{token embedding of } z_t^l & l \notin \gM_t \quad \text{(fixed, already decoded)}
        \end{cases}$ \\
        sg stands for {\em stop gradient}.
        \STATE Update $\vq^l$ via gradient ascent on logits w.r.t $R(\ve_\text{EntRGi})$ for $l \in \gM_t$.
    \ENDFOR
    \STATE Unmask tokens $z^l_{t-1} \sim \vq^l$ for $l\in \gU(\vq^{\gM_t})$
    \STATE Copy over all other tokens (masked or unmasked), i.e. $z^l_{t-1} = z^l_t$ for all $l\notin \mathcal{U}(\vq^{\gM_t})$
\ENDFOR
\STATE \textbf{return} reward-guided string $z_0$
\end{algorithmic}
\end{algorithm}

\begin{algorithm}[h]
\caption{RGRL: Reward Guided Reinforcement Learning for dLLMs}
\label{alg:EntRGi-train}
\begin{algorithmic}[1]
\REQUIRE dLLM $\vp_\theta$, reward model $R$, completions per prompt $N$, Dataset $\mathcal{D}$
\FOR{each training step}
    \STATE Sample prompt $c \sim \mathcal{D}$
    \FOR{$n = 1, \dots, N$}
        \STATE Sample completion $y_n$ via~\autoref{alg:EntRGi} using $\vp_\theta$ and $R$
    \ENDFOR
    \STATE Estimate $\log \vp_\theta(y_n \mid c)$, following \citet{d1}
    \STATE Update $\theta$ to minimize $\mathcal{L}(\theta) = {-}\frac{1}{N}\sum_{n=1}^N \log \vp_\theta(y_n \mid c)$
\ENDFOR
\end{algorithmic}
\end{algorithm}

\autoref{alg:EntRGi} summarizes the full inference-time procedure. At each
denoising step $t$, we run $M$ inner updates on $\vq^l$ at
currently masked positions, with gradients flowing through the entropy-weighted embedding ${\ve}_\text{EntRGi}^l$ defined in \autoref{eq:ehat}. The updated logits define a reward-tilted distribution from which we sample the tokens to commit; remaining positions stay masked and are revisited at later steps.

\textbf{RGRL: Reward Guidance for Reinforcement Learning.}
We further demonstrate how EntRGi and APS \citep{aps} can be applied as post-training algorithms by self-distillation of reward-guided completions. As shown in \autoref{alg:EntRGi-train}, at each step we draw a prompt $c \sim \mathcal{D}$, generate $N$ completions $\{y_n\}_{n=1}^N$ via
\autoref{alg:EntRGi} under the current parameters $\theta$, and update
$\theta$ to increase the likelihood of reward-guided completions. Depending on whether we use EntRGi or APS ($w^l=1$) for generations, we refer to these algorithms RGRL-EntRGi and RGRL-APS, respectively. 

\textbf{Mismatched tokenizers.} The formulation above assumes that the dLLM and the reward model $R$ share the same vocabulary $\gV$, enabling us to compute the expected embedding ${\ve}_\text{soft}^l = \sum_{k=1}^K \vq^l_k\,\mE^R_k$. In practice, this assumption may fail: reward models are sometimes fine-tuned from base models with a different tokenizer, yielding vocabularies $\gV^P$ (dLLM) and $\gV^R$ (reward) that only partially overlap. Let $\gV^{P \cap R} = {k \in \gV^P : k \in \gV^R}$ denote the vocabulary intersection. For tokens $k \in \gV^{P \cap R}$, the corresponding reward embedding $\mE^R_k$ is well-defined. For tokens $k \in \gV^P \setminus \gV^R$ that are absent from the reward vocabulary, we set $\mE^R_k = \mathbf{0}$. Hence, only matched tokens contribute to the gradient.  

\section{Experiments}
\label{sec:exps}

\textbf{Models.}
We use \textit{Dream-v0-Instruct-7B}\footnote{\href{https://huggingface.co/Dream-org/Dream-v0-Instruct-7B}{Dream-org/Dream-v0-Instruct-7B}}~\citep{dream} and \textit{LLaDA-8B-Instruct}\footnote{\href{https://huggingface.co/GSAI-ML/LLaDA-8B-Instruct}{GSAI-ML/LLaDA-8B-Instruct}} as the base dLLMs in all experiments. As reward models, we adopt the Skywork family~\citep{skywork}, which demonstrates strong performance across diverse domains including safety, factuality, and helpfulness~\citep{rewardbench2}. Our experiments encompass 4 publicly available reward model sizes spanning two different model families -- more details are provided in~\autoref{sec:appdx_exp_setup}.

\begin{table*}[h]
\centering
\footnotesize
\addtolength{\tabcolsep}{-2pt}
\caption{Performance of \textit{Dream-v0-7B-Instruct} on Reward-Bench-2 \citep{rewardbench2}, JudgeBench \citep{judgebench}, and RM-Bench \citep{rmbench} with \textit{Skywork-Reward-v2-Qwen3-1.7B} as the reward model. EntRGi outperforms APS \cite{aps} on majority of tasks, and shows stronger overall performance at higher temperatures ($\tau$=0.7).}
\label{tab:main_results}
\begin{tabular}{lccccccccc}
\toprule
\multirow{2}{*}{\textbf{Method}} & \multicolumn{3}{c}{Reward-Bench-2} & \multicolumn{3}{c}{JudgeBench} & \multicolumn{3}{c}{RM-Bench} \\ \cmidrule(lr){2-4} \cmidrule(lr){5-7} \cmidrule(lr){8-10}
 & \textbf{Top@1} & \textbf{Avg@4} & \textbf{LMUnit} & \textbf{Top@1} & \textbf{Avg@4} & \textbf{LMUnit} & \textbf{Top@1} & \textbf{Avg@4} & \textbf{LMUnit} \\ \midrule
\multicolumn{10}{c}{Temperature ($\tau=0.1$)} \\ \midrule
BoN & 0.18\err{0.22} & 0.05\err{0.23} & 3.74\err{0.04} & 0.00\err{0.15} & {\ul -0.07}\err{0.16} & 3.75\err{0.03} & 3.05\err{0.05} & \textbf{3.02}\err{0.05} & 3.93\err{0.01} \\
Expectation & 2.19\err{0.19} & \textbf{1.62}\err{0.17} & 4.12\err{0.03} & 0.68\err{0.19} & \textbf{-0.06}\err{0.21} & 3.81\err{0.02} & 3.33\err{0.20} & 2.59\err{0.12} & 3.89\err{0.04} \\
APS & {\ul 2.95}\err{0.21} & 1.47\err{0.20} & {\ul 4.19}\err{0.01} & \textbf{1.67}\err{0.11} & -0.17\err{0.14} & {\ul 3.89}\err{0.03} & {\ul 4.72}\err{0.13} & 2.46\err{0.17} & {\ul 4.01}\err{0.03} \\
\rowcolor{highlight} EntRGi & \textbf{3.07}\err{0.22} & \textbf{1.62}\err{0.18} & \textbf{4.22}\err{0.02} & \textbf{1.73}\err{0.14} & -0.11\err{0.18} & \textbf{3.94}\err{0.01} & \textbf{4.90}\err{0.13} & {\ul 2.75}\err{0.14} & \textbf{4.06}\err{0.01} \\ \midrule
\multicolumn{10}{c}{Temperature ($\tau=0.7$)} \\ \midrule
BoN & 2.99\err{0.23} & 1.38\err{0.29} & 4.15\err{0.02} & 1.65\err{0.18} & -0.84\err{0.16} & 3.91\err{0.02} & 5.11\err{0.20} & 2.98\err{0.15} & {\ul 4.02}\err{0.03} \\
Expectation & \textbf{3.95}\err{0.28} & \textbf{2.23}\err{0.24} & {\ul 4.22}\err{0.02} & {\ul 2.30}\err{0.08} & \textbf{0.13}\err{0.07} & \textbf{3.97}\err{0.01} & {\ul 5.45}\err{0.16} & {\ul 3.29}\err{0.13} & {\ul 4.02}\err{0.03} \\
APS & {\ul 3.62}\err{0.27} & 1.80\err{0.24} & {\ul 4.22}\err{0.02} & 1.87\err{0.14} & -0.63\err{0.10} & 3.93\err{0.02} & 5.11\err{0.14} & 2.66\err{0.15} & 4.00\err{0.02} \\
\rowcolor{highlight} EntRGi & \textbf{3.91}\err{0.30} & \textbf{2.20}\err{0.26} & \textbf{4.25}\err{0.02} & \textbf{2.44}\err{0.06} & {\ul 0.02}\err{0.10} & \textbf{3.98}\err{0.02} & \textbf{5.70}\err{0.12} & \textbf{3.41}\err{0.14} & \textbf{4.04}\err{0.01} \\ \bottomrule
\end{tabular}
\end{table*}
\subsection{Test-Time Adaptation}
\label{sec:eval-results}

\textit{Datasets.} We source prompts from three benchmarking suites: Reward-Bench-2 \citep{rewardbench2}, RM-Bench \citep{rmbench}, and JudgeBench \citep{judgebench}. These datasets contain prompts that measure multiple fine-grained chatbot abilities, such as precise instruction following, safety, factuality, and knowledge.

\textit{Metrics.}
We evaluate each final, discretized completion using the reward model. Specifically, we report the maximum reward across samples (Top@1) and the average reward across all $N$ trajectories per prompt (Avg@$N$), with $N=4$ unless stated otherwise. Top@1 measures the best achievable outcome, while Avg@$N$ reflects overall generation quality. To detect possible reward hacking or overoptimization \citep{gao2022scalinglawsrewardmodel}, we additionally use \textit{LMUnit-Qwen2.5-72B}~\citep{lmunit} as an external judge. More details are provided in~\appref{subsec:appdx_model_inputs}. We qualitatively analyze generations in~\appref{sec:appdx_qa}.

\textit{Baselines.}
We consider Best-of-$N$ (BoN) as widely-used gradient-free reference point~\citep{rewardbench2,rmbench}. BoN generates $N$ independent trajectories and selects the highest-scoring sample. Among gradient-based baselines, we evaluate Expectation which directly feeds a continuous convex combination of token probabilities and reward-model embeddings~\citep{tess-2,g2d2}. Finally, we compare against APS~\citep{aps}, a strong prior method that updates logits at each denoising step by feeding discretized tokens to the reward model via the straight-through estimator (STE)~\citep{ste,gumbel-soft}. All gradient methods incur computational cost due to reward model gradients; we analyze compute–performance trade-offs in~\appref{subsec_appdx:throughput}.

\takeaway{Gradient-based methods outperform BoN.}
As shown in \autoref{tab:main_results}, all gradient-based methods consistently outperform Best-of-N (BoN) across all benchmarks. Gradient-based guidance can be viewed as performing directed search in the continuous space spanned by token embeddings, whereas BoN relies on zeroth-order sampling by selecting from a finite set of randomly generated trajectories. 

\takeaway{Expectation at low-entropy positions provide consistent improvements.} EntRGi achieves a relative improvement of approximately 33\% over APS in reward-model-judged output quality. EntRGi additionally improves the LMUnit score on RewardBench-2 from 4.19 (APS) to 4.22, and on RM-Bench from 4.01 to 4.06, while also achieving higher Top@1 reward across all tasks. EntRGi further improves at higher sampling temperature ($\tau$=0.7), achieving the strongest results, while APS noticeably degrades.

\begin{table*}[h]
\centering
\small
\addtolength{\tabcolsep}{0pt}
\caption{Performance of \textit{LLaDA-8B-Instruct} under mismatched tokenizer (45\% mismatch with Llama, 55\% with Qwen) on Reward-Bench-2 \citep{rewardbench2}, JudgeBench \citep{judgebench}, and RM-Bench \citep{rmbench}. Sampling temperature $\tau=0.7$.}
\label{tab:tokenizer_results_llada_appdx}
\begin{tabular}{lcccccc}
\toprule
\multirow{2}{*}{\textbf{Method}} & \multicolumn{2}{c}{Reward-Bench-2} & \multicolumn{2}{c}{JudgeBench} & \multicolumn{2}{c}{RM-Bench} \\ \cmidrule(lr){2-3} \cmidrule(lr){4-5} \cmidrule(lr){6-7}
 & \textbf{Top@1} & \textbf{Avg@4} & \textbf{Top@1} & \textbf{Avg@4} & \textbf{Top@1} & \textbf{Avg@4} \\
\midrule
\multicolumn{7}{c}{\textit{Skywork-Reward-V2-Llama-3.2-1B}} \\
\midrule
BoN            & 5.34\err{0.38}        & 3.73\err{0.31} & 5.92\err{0.07}        & 4.42\err{0.08} & 9.05\err{0.15}        & 7.28\err{0.16} \\
 Expectation & 5.95\err{0.31}        & 4.26\err{0.36} & \textbf{6.45}\err{0.13}        & \textbf{4.79}\err{0.09} & {\ul 9.52}\err{0.21}        & \textbf{7.47}\err{0.15} \\
 APS         & 6.32\err{0.48}        & {\ul 4.28}\err{0.36} & 6.33\err{0.07}        & 4.49\err{0.05} & 9.19\err{0.19}        & 6.93\err{0.17} \\
\rowcolor{highlight}  EntRGi & \textbf{6.40}\err{0.33} & \textbf{4.50}\err{0.31} & \textbf{6.51}\err{0.11}        & \textbf{4.73}\err{0.10} & \textbf{9.74}\err{0.09}        & \textbf{7.44}\err{0.13} \\
\midrule
\multicolumn{7}{c}{\textit{Skywork-Reward-V2-Qwen-3-0.6B}} \\
\midrule
BoN            & 1.77\err{0.20}        & 0.52\err{0.21} & 2.29\err{0.14}        & 0.83\err{0.15} & 3.90\err{0.15}        & 2.42\err{0.10} \\
 Expectation & 2.72\err{0.22}        & \textbf{1.27}\err{0.26} & {\ul 2.87}\err{0.08}        & {\ul 1.15}\err{0.12} & {\ul 4.64}\err{0.15}        & {\ul 2.82}\err{0.17} \\
 APS         & 2.35\err{0.19}        & 0.85\err{0.23} & {\ul 2.72}\err{0.12}        & 0.79\err{0.10} & 4.52\err{0.15}        & 2.46\err{0.18} \\
\rowcolor{highlight}  EntRGi & \textbf{2.80}\err{0.20}        & \textbf{1.31}\err{0.24} & \textbf{3.19}\err{0.11} & \textbf{1.22}\err{0.12} & \textbf{4.85}\err{0.07}        & \textbf{2.88}\err{0.06} \\
\bottomrule
\end{tabular}
\end{table*}

\takeaway{STE is critical at high-entropy positions.}
Removing STE at high-entropy positions ($w=0$) reduces EntRGi to the Expectation baseline. As shown in \autoref{tab:main_results}, EntRGi consistently outperforms Expectation, highlighting the importance of STE in these regimes.
At the beginning of the denoising process ($t=T$), the per-token entropy is typically high at most positions due to limited contextual information. APS treats all positions uniformly and applies the STE regardless of entropy, which incurs large approximation error $\mathcal{E}^l$ at positions where soft representations would be more appropriate.

In contrast, EntRGi adaptively selects soft representations at positions $l$, which reduces the approximation error. To receive reliable gradients at $l$, the reward model must see realistic hard tokens at the remaining high-entropy positions $\{1,\ldots,l-1,l+1,\ldots,L\}$ because it requires an entire sequence to compute the score. EntRGi automatically adjusts hardness via STE, as ${\ve}_\text{EntRGi}^l \rightarrow {\ve}_\text{hard}^l$ when $w^l \rightarrow 1$, justifying why STE is critical in this regime.

\takeaway{EntRGi demonstrates performance gains in mismatched tokenizer settings.} We apply EntRGi to \textit{LLaDA-8B-Instruct} \citep{llada}, whose tokenizer overlap is 45\% and 55\% with \textit{Qwen3-0.6B} and \textit{Llama-3.2-1B}, respectively. Following the procedure in \autoref{sec:method}, non-overlapping tokens receive no gradients. As shown in \autoref{tab:tokenizer_results_llada_appdx}, the trends mirror those observed on Dream: gradient-based methods outperform BoN (which does not need to handle mismatch), and EntRGi performs best. We observe similar trends in post-training, as shown in a later section.

\subsection{Reward Guided Post-Training}

\textit{Datasets.} For post-training experiments, we source prompts from the WildChat-IF subset of the T{\"u}lu SFT mixture~\citep{zhao2024wildchat1mchatgptinteraction, bhaskar2025languagemodelsthinkchat}, and lmsys-chat-1M~\citep{zheng2024lmsyschat1mlargescalerealworldllm}. Conditioned on these prompts, we apply \autoref{alg:EntRGi-train} to update the dLLM.

\textit{Metrics.} We follow \citep{d1, shao2024deepseekmathpushinglimitsmathematical} and report the average reward over the $N$ parallel trajectories. We use \textit{Skywork-Reward-V2-Qwen3-0.6B} \citep{skywork} as the reward model.

\textit{Baselines.} We compare against diffu-GRPO \citep{d1}, a widely adopted RL algorithm for dLLMs. diffu-GRPO does not compute any reward model gradients, rather relies on policy-gradient based updates using scalar rewards from sampled trajectories. We instantiate our RGRL recipe with two choices of reward guidance: APS (RGRL-APS) and EntRGi (RGRL-EntRGi). Further details are provided in~\autoref{sec:appdx_exp_setup}.

\begin{figure*}[h]
  \centering
  \includegraphics[width=\linewidth]{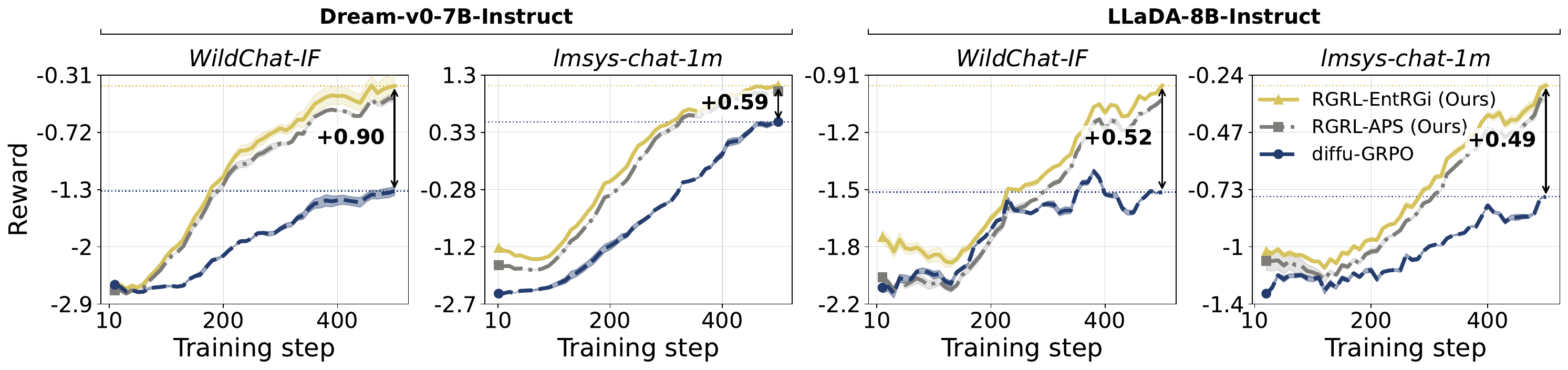}
  \caption{Training curves on WildChat-IF and  \citep{bhaskar2025languagemodelsthinkchat, zhao2024wildchat1mchatgptinteraction} and lmsys-chat-1m \citep{zheng2024lmsyschat1mlargescalerealworldllm} with reward as per \textit{Skywork-Reward-V2-Qwen3-0.6B}. \textit{LLaDA-8B-Instruct} is tokenizer-mismatched.}
  \label{fig:rl_experiment}
\end{figure*}
\takeaway{RGRL improves sample efficiency.}
As shown in~\autoref{fig:rl_experiment}, RGRL demonstrates consistent gains over diffu-GRPO~\citep{d1} when controlling for the number of training steps (or examples seen). We attribute these improvements to the use of dense reward signals via gradient feedback from the reward model, consistent with prior findings in image diffusion~\citep{cg, prabhudesai2024videodiffusionalignmentreward}. On WildChat-IF with Dream, we observe a relative improvement of up to 70\% (+0.90 absolute improvement).
In terms of compute efficiency, we observe faster convergence in terms of wall-clock time in 1 out of 4 settings (see \autoref{fig:rl_experiment_appdx} in \appref{sec:appdx_exp_setup}). This is expected, as differentiating through the reward model introduces additional computational overhead. RGRL-EntRGi yields the highest sustained gains across all settings.

\takeaway{RGRL demonstrates gains under tokenizer mismatch.}
Consistent with our test-time observations, RGRL variants continue to outperform the respective baselines on LLaDA despite tokenizer mismatch. However, the absolute gains are smaller in this setting, reaching up to +0.52 on WildChat-IF.

\takeaway{Increased gains on harder datasets.}
Across both Dream and LLaDA, we observe that RGRL's improvements scale inversely with the initial reward: gains are largest on the WildChat-IF dataset, smaller on lmsys-chat-1m, and smallest on Magpie-Ultra (see \autoref{fig:magpie_rl_experiment} in the Appendix), where initial rewards are already nearly positive. 

\begin{figure*}[h]
  \centering
  \includegraphics[width=\linewidth]{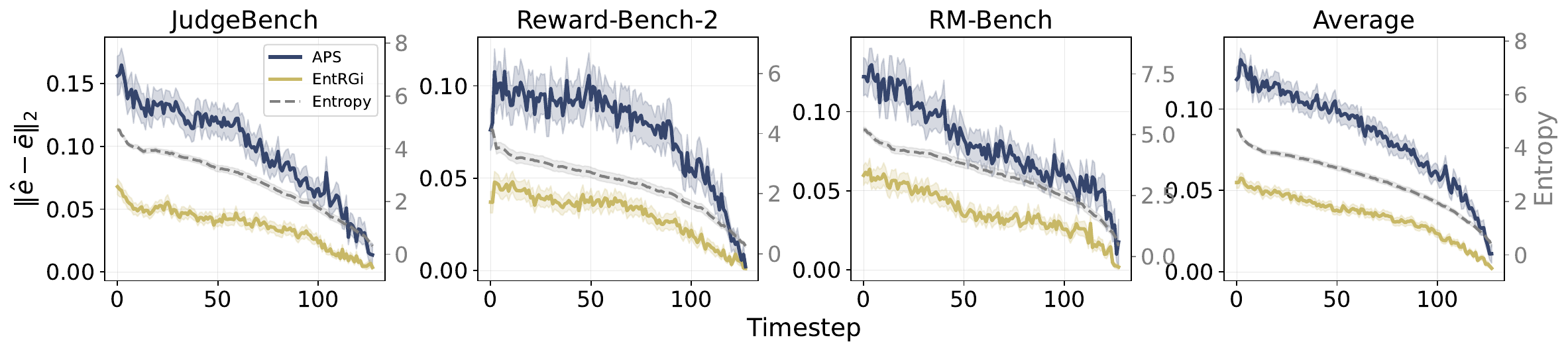}
  \caption{Average L2-norm between the soft embedding $\bar{\ve}=\ve_\text{soft}$ and the reward model input $\hat{\ve}=\ve_\text{EntRGi/APS}$ vs. decoding timestep, along with average entropy. The maximum possible entropy is $\log K \approx 11$. EntRGi reduces early-step approximation error compared to APS by upweighting the continuous relaxation on tokens with relatively low entropy in the predicted sequence.}
  \label{fig:l2_vs_error}
\end{figure*}

\subsection{Analysis}

\takeaway{EntRGi reduces approximation error during early denoising steps.}
To further analyze EntRGi’s behavior over the denoising trajectory, we examine the L2 discrepancy between the reward model input $\hat{\ve}$ (which can be either $\ve_\text{EntRGi}$ or $\ve_\text{APS}$) and the soft embedding $\bar{\ve}=\ve_\text{soft}$ across timesteps. \autoref{fig:l2_vs_error} reports this error averaged over sequence length $L=128$ and 32 prompts\footnote{\autoref{fig:l2_vs_error} aggregates over tokens; we provide a finer-grained per-token entropy–error histogram in \appref{subsec:appdx_error_histogram}.}. At the initial denoising step ($t=T$), all tokens contribute to the approximation error, since the sequence is fully masked. As denoising progresses and tokens become increasingly determined, fewer positions contribute, leading to a natural decay in error as $t \rightarrow 0$.

In moderate- to high-entropy regimes (entropy $\approx 4$--$6$), APS often samples discrete tokens whose embeddings $\ve_\text{hard}^l$ deviate substantially from ${\ve}_\text{soft}^l$, resulting in large approximation error in early decoding. In contrast, EntRGi leverages token-level entropy to adaptively weight the soft embedding $\ve_\text{soft}^l$, reducing this discrepancy by trading off vocabulary error against reward-model reliability. 
As denoising progresses, the approximation error of both methods converges to zero.

\begin{figure*}[h]
  \centering
  \includegraphics[width=\linewidth]{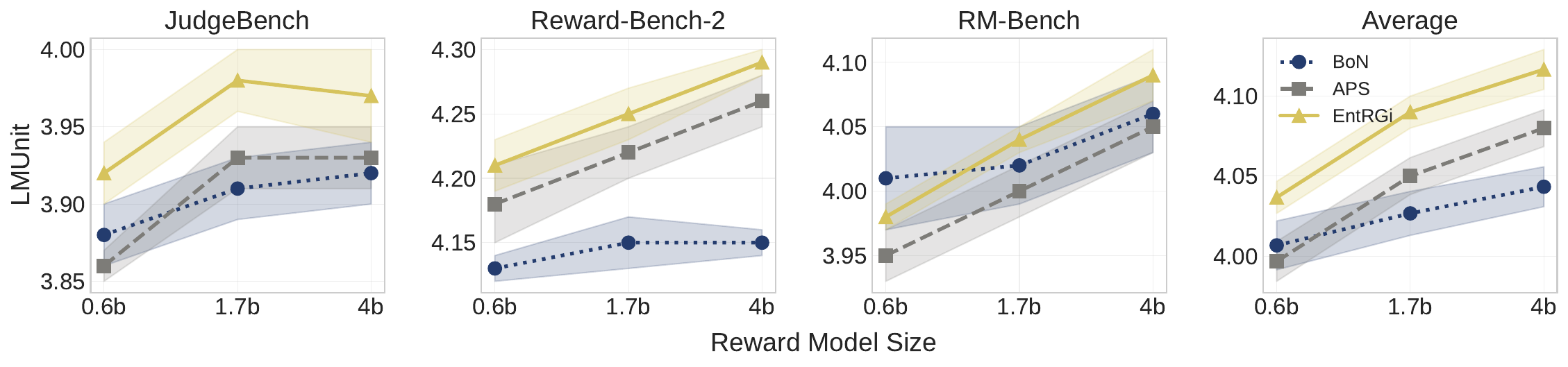}
  \caption{LMUnit score with increasing reward model size across Reward-Bench-2 \citep{rewardbench2}, RM-Bench \citep{rmbench}, and JudgeBench \citep{judgebench}, for $M=3$ and $\tau=0.7$. Increasing reward model size leads to improved performance. We observe similar trends for other metrics (Top@1, Avg@4) in \appref{sec:appdx_rm_size}.}
  \label{fig:scaling_rm_size}
\end{figure*}
\takeaway{EntRGi benefits from increasing reward model size.} In \autoref{fig:scaling_rm_size}, we study the effect of reward model size, ranging from 0.6B to 4B parameters. Across all three datasets, increasing reward model size leads to consistent improvements in scores as measured by LMUnit for all methods. For instance, APS improves from an average LMUnit score of 4.00 at 0.6B to 4.08 at 4B, while EntRGi improves from 4.04 to 4.12 over the same range. At each reward model size, EntRGi achieves better score, outperforming APS across all datasets. These results show that larger reward models improve overall performance, while EntRGi maintains its advantage across reward model scales.

\begin{figure*}
  \centering
  \includegraphics[width=\linewidth]{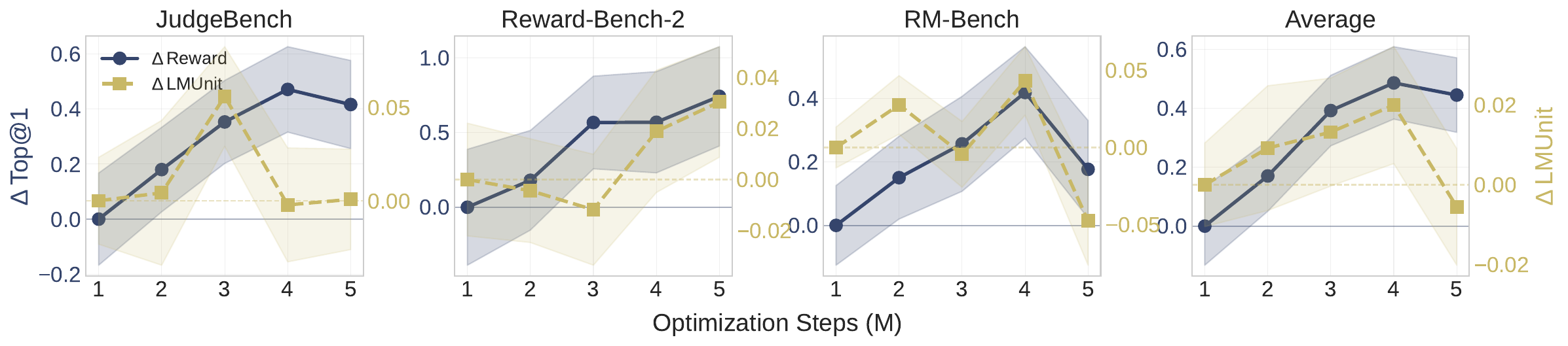}
  \caption{Change in Top@1 accuracy and LMUnit score relative to $M=1$ as reward model gradient steps $M$ increase for EntRGi. Results are averaged over 3 reward model sizes (0.6B, 1.7B, 4B). Optimal $M$ is dataset-dependent (our experiments use $M=3$ for all datasets). LMUnit collapses beyond $M=4$, indicating overoptimization. Raw scores are reported in \appref{sec:appdx_m_scaling}.}
  \label{fig:scaling_m}
\end{figure*}
\takeaway{Increasing reward model gradient steps improves performance but risks over-optimization.}
In \autoref{fig:scaling_m}, we analyze the effect of increasing the number of optimization steps \(M\).
Increasing \(M\) from 1 to approximately 3--4 leads to consistent improvements in both reward and LMUnit scores on JudgeBench and RM-Bench, after which performance begins to degrade.
On Reward-Bench-2, reward scores roughly improve up to \(M = 5\). Overall, \(M = 3\)--4 represents a reliable operating range in which both reward and LMUnit scores improve consistently across benchmarks.
These observations suggest that (i) the optimal number of optimization steps varies across datasets, motivating further investigation in future work, and (ii) drastically increasing \(M\) may lead to ``reward hacking'' or over-optimization~\citep{gao2022scalinglawsrewardmodel, moskovitz2023confrontingrewardmodeloveroptimization}.

\section{Conclusion}
\label{sec:conclusion}

We introduced \textbf{EntRGi}, a reward guidance method for discrete diffusion language models that dynamically interpolates between continuous relaxations and hard token embeddings based on the model's predictive entropy. This simple mechanism addresses the fundamental tension between gradient accuracy and reward-model reliability: trusting soft embeddings when the model is confident, and reverting to discrete tokens when uncertainty is high. We then presented \textbf{RGRL}, the first study of post-training dLLMs by leveraging reward gradients during rollouts, finding that it surpasses widely-adopted scalar-reward RL. Together, these results suggest that the model's uncertainty is a useful signal for regulating how reward feedback is incorporated, with EntRGi enabling improved inference-time steering and RGRL extending these benefits to post-training of dLLMs.

\textbf{Limitations and Future Work.} Like all gradient-based reward guidance methods, EntRGi/RGRL require a differentiable reward model and incur additional compute from back-propagating through it, which can offset wall-clock gains in some settings. Promising directions to address these and extend the method include (i) selectively applying reward-gradient feedback only at the most informative denoising steps to better trade off sample efficiency against compute, (ii) extending to multi-objective reward composition, (iii) developing more principled treatments of tokenizer mismatch, and (iv) leveraging differentiable-reward models for dense feedback in RLVR-style methods.

\textbf{Broader Impacts.}
This paper presents work whose goal is to improve the alignment of discrete diffusion language models.  It inherits risks common to reward-guided systems, including potential reward hacking and misalignment between proxy rewards and true human preferences. Additionally, enhanced controllability could be misused to generate targeted harmful content. We recommend precautions and auxiliary quality checks when deploying such methods.

\section*{Acknowledgments}
This research has been supported by NSF Grants 2217069, 2019844 and 2112471, the UT Austin Machine
Learning Lab, and computing support on the Vista GPU Cluster through the Center for Generative AI
(CGAI) and the Texas Advanced Computing Center (TACC) at UT Austin.

\printbibliography

@techreport{geminidiffusion,
  author      = {{DeepMind}},
  title       = {Gemini Diffusion},
  institution = {DeepMind},
  year        = {2025},
  url         = {https://deepmind.google/models/gemini-diffusion/},
  note        = {Accessed: 2026-01-24}
}

@article{aps,
  title={Test-Time Anchoring for Discrete Diffusion Posterior Sampling},
  author={Rout, Litu and Lugmayr, Andreas and Jafarian, Yasamin and Varadharajan, Srivatsan and Caramanis, Constantine and Shakkottai, Sanjay and Kemelmacher-Shlizerman, Ira},
  journal={arXiv preprint arXiv:2510.02291},
  year={2025},
  url={https://arxiv.org/pdf/2510.02291}
}

@inproceedings{
rfi,
title={Semantic Image Inversion and Editing using Rectified Stochastic Differential Equations},
author={Litu Rout and Yujia Chen and Nataniel Ruiz and Constantine Caramanis and Sanjay Shakkottai and Wen-Sheng Chu},
booktitle={The Thirteenth International Conference on Learning Representations},
year={2025},
url={https://openreview.net/forum?id=Hu0FSOSEyS}
}

@inproceedings{
gumbel-soft,
title={Categorical Reparameterization with Gumbel-Softmax},
author={Eric Jang and Shixiang Gu and Ben Poole},
booktitle={International Conference on Learning Representations},
year={2017},
url={https://openreview.net/forum?id=rkE3y85ee}
}

@article{svdd,
  title={Derivative-free guidance in continuous and discrete diffusion models with soft value-based decoding},
  author={Li, Xiner and Zhao, Yulai and Wang, Chenyu and Scalia, Gabriele and Eraslan, Gokcen and Nair, Surag and Biancalani, Tommaso and Ji, Shuiwang and Regev, Aviv and Levine, Sergey and others},
  journal={arXiv preprint arXiv:2408.08252},
  year={2024},
  url={https://arxiv.org/pdf/2408.08252}
}

@InProceedings{p2l,
  title = 	 {Prompt-tuning Latent Diffusion Models for Inverse Problems},
  author =       {Chung, Hyungjin and Ye, Jong Chul and Milanfar, Peyman and Delbracio, Mauricio},
  booktitle = 	 {Proceedings of the 41st International Conference on Machine Learning},
  pages = 	 {8941--8967},
  year = 	 {2024},
  volume = 	 {235},
  series = 	 {Proceedings of Machine Learning Research},
  publisher =    {PMLR},
  url = 	 {https://proceedings.mlr.press/v235/chung24b.html}
  }

@inproceedings{
psld,
title={Solving Inverse Problems Provably via Posterior Sampling with Latent Diffusion Models},
author={Rout, Litu and Raoof, Negin and Daras, Giannis and Caramanis, Constantine and Dimakis, Alexandros G and Shakkottai, Sanjay},
booktitle={Thirty-seventh Conference on Neural Information Processing Systems},
year={2023},
url={https://openreview.net/forum?id=XKBFdYwfRo}
}

@article{sgdd,
  title={Split gibbs discrete diffusion posterior sampling},
  author={Chu, Wenda and Wu, Zihui and Chen, Yifan and Song, Yang and Yue, Yisong},
  journal={arXiv preprint arXiv:2503.01161},
  year={2025},
  url={https://arxiv.org/pdf/2503.01161}
}

@article{g2d2,
  title={G2D2: Gradient-guided Discrete Diffusion for image inverse problem solving},
  author={Murata, Naoki and Lai, Chieh-Hsin and Takida, Yuhta and Uesaka, Toshimitsu and Nguyen, Bac and Ermon, Stefano and Mitsufuji, Yuki},
  journal={arXiv preprint arXiv:2410.14710v1},
  year={2024},
  url={https://arxiv.org/abs/2410.14710v1}
}

@article{llada,
  title={Large Language Diffusion Models},
  author={Nie, Shen and Zhu, Fengqi and You, Zebin and Zhang, Xiaolu and Ou, Jingyang and Hu, Jun and Zhou, Jun and Lin, Yankai and Wen, Ji-Rong and Li, Chongxuan},
  journal={arXiv preprint arXiv:2502.09992},
  year={2025},
  url={https://arxiv.org/pdf/2502.09992}
}

@inproceedings{
mdlm,
title={Simple and Effective Masked Diffusion Language Models},
author={Subham Sekhar Sahoo and Marianne Arriola and Aaron Gokaslan and Edgar Mariano Marroquin and Alexander M Rush and Yair Schiff and Justin T Chiu and Volodymyr Kuleshov},
booktitle={The Thirty-eighth Annual Conference on Neural Information Processing Systems},
year={2024},
url={https://openreview.net/forum?id=L4uaAR4ArM}
}

@inproceedings{
d3pm,
title={Structured Denoising Diffusion Models in Discrete State-Spaces},
author={Jacob Austin and Daniel D. Johnson and Jonathan Ho and Daniel Tarlow and Rianne van den Berg},
booktitle={Advances in Neural Information Processing Systems},
editor={A. Beygelzimer and Y. Dauphin and P. Liang and J. Wortman Vaughan},
year={2021},
url={https://openreview.net/forum?id=h7-XixPCAL}
}

@inproceedings{
sedd,
title={Discrete Diffusion Modeling by Estimating the Ratios of the Data Distribution},
author={Aaron Lou and Chenlin Meng and Stefano Ermon},
booktitle={Forty-first International Conference on Machine Learning},
year={2024},
url={https://openreview.net/forum?id=CNicRIVIPA}
}

@inproceedings{
md4,
title={Simplified and Generalized Masked Diffusion for Discrete Data},
author={Jiaxin Shi and Kehang Han and Zhe Wang and Arnaud Doucet and Michalis Titsias},
booktitle={The Thirty-eighth Annual Conference on Neural Information Processing Systems},
year={2024},
url={https://openreview.net/forum?id=xcqSOfHt4g}
}

@inproceedings{
rbm,
title={{RB}-Modulation: Training-Free Stylization using Reference-Based Modulation},
author={Litu Rout and Yujia Chen and Nataniel Ruiz and Abhishek Kumar and Constantine Caramanis and Sanjay Shakkottai and Wen-Sheng Chu},
booktitle={The Thirteenth International Conference on Learning Representations},
year={2025},
url={https://openreview.net/forum?id=bnINPG5A32}
}

@INPROCEEDINGS{stsl,
  author={Rout, Litu and Chen, Yujia and Kumar, Abhishek and Caramanis, Constantine and Shakkottai, Sanjay and Chu, Wen-Sheng},
  booktitle={2024 IEEE/CVF Conference on Computer Vision and Pattern Recognition}, 
  title={Beyond First-Order Tweedie: Solving Inverse Problems using Latent Diffusion}, 
  year={2024},
  url={https://arxiv.org/pdf/2312.00852}
  }

@article{stylealigned,
  title={Style aligned image generation via shared attention},
  author={Hertz, Amir and Voynov, Andrey and Fruchter, Shlomi and Cohen-Or, Daniel},
  journal={arXiv preprint arXiv:2312.02133},
  year={2023}
}

@inproceedings{dps,
title={Diffusion Posterior Sampling for General Noisy Inverse Problems},
author={Hyungjin Chung and Jeongsol Kim and Michael Thompson Mccann and Marc Louis Klasky and Jong Chul Ye},
booktitle={The Eleventh International Conference on Learning Representations },
year={2023},
url={https://openreview.net/forum?id=OnD9zGAGT0k}
}

@article{cg,
  title={Diffusion models beat gans on image synthesis},
  author={Dhariwal, Prafulla and Nichol, Alexander},
  journal={Advances in Neural Information Processing Systems},
  volume={34},
  pages={8780--8794},
  year={2021},
  url={https://proceedings.neurips.cc/paper_files/paper/2021/file/49ad23d1ec9fa4bd8d77d02681df5cfa-Paper.pdf}
}

@misc{rewardbench2,
      title={RewardBench 2: Advancing Reward Model Evaluation}, 
      author={Saumya Malik and Valentina Pyatkin and Sander Land and Jacob Morrison and Noah A. Smith and Hannaneh Hajishirzi and Nathan Lambert},
      year={2025},
      eprint={2506.01937},
      archivePrefix={arXiv},
      primaryClass={cs.CL},
      url={https://arxiv.org/abs/2506.01937}, 
}

@misc{judgebench,
      title={JudgeBench: A Benchmark for Evaluating LLM-based Judges}, 
      author={Sijun Tan and Siyuan Zhuang and Kyle Montgomery and William Y. Tang and Alejandro Cuadron and Chenguang Wang and Raluca Ada Popa and Ion Stoica},
      year={2025},
      eprint={2410.12784},
      archivePrefix={arXiv},
      primaryClass={cs.AI},
      url={https://arxiv.org/abs/2410.12784}, 
}

@misc{rmbench,
      title={RM-Bench: Benchmarking Reward Models of Language Models with Subtlety and Style}, 
      author={Yantao Liu and Zijun Yao and Rui Min and Yixin Cao and Lei Hou and Juanzi Li},
      year={2024},
      eprint={2410.16184},
      archivePrefix={arXiv},
      primaryClass={cs.CL},
      url={https://arxiv.org/abs/2410.16184}, 
}

@misc{pg-dlm,
      title={Inference-Time Scaling of Diffusion Language Models with Particle Gibbs Sampling}, 
      author={Meihua Dang and Jiaqi Han and Minkai Xu and Kai Xu and Akash Srivastava and Stefano Ermon},
      year={2025},
      eprint={2507.08390},
      archivePrefix={arXiv},
      primaryClass={cs.LG},
      url={https://arxiv.org/abs/2507.08390}, 
}

@misc{dream,
      title={Dream 7B: Diffusion Large Language Models}, 
      author={Jiacheng Ye and Zhihui Xie and Lin Zheng and Jiahui Gao and Zirui Wu and Xin Jiang and Zhenguo Li and Lingpeng Kong},
      year={2025},
      eprint={2508.15487},
      archivePrefix={arXiv},
      primaryClass={cs.CL},
      url={https://arxiv.org/abs/2508.15487}, 
}

@article{skywork,
  title={Skywork-Reward-V2: Scaling Preference Data Curation via Human-AI Synergy},
  author={Liu, Chris Yuhao and Zeng, Liang and Xiao, Yuzhen and He, Jujie and Liu, Jiacai and Wang, Chaojie and Yan, Rui and Shen, Wei and Zhang, Fuxiang and Xu, Jiacheng and others},
  journal={arXiv preprint arXiv:2507.01352},
  year={2025}
}

@misc{lmunit,
      title={LMUnit: Fine-grained Evaluation with Natural Language Unit Tests}, 
      author={Jon Saad-Falcon* and Rajan Vivek* and William Berrios* and Nandita Shankar Naik and Matija Franklin and Bertie Vidgen and Amanpreet Singh and Douwe Kiela and Shikib Mehri},
      year={2024},
      eprint={2412.13091},
      archivePrefix={arXiv},
      primaryClass={cs.CL},
      url={https://arxiv.org/abs/2412.13091},
      note={*Equal contribution}
}

@inproceedings{tess-2,
    title = "{TESS} 2: A Large-Scale Generalist Diffusion Language Model",
    author = "Tae, Jaesung  and
      Ivison, Hamish  and
      Kumar, Sachin  and
      Cohan, Arman",
    editor = "Che, Wanxiang  and
      Nabende, Joyce  and
      Shutova, Ekaterina  and
      Pilehvar, Mohammad Taher",
    booktitle = "Proceedings of the 63rd Annual Meeting of the Association for Computational Linguistics (Volume 1: Long Papers)",
    month = jul,
    year = "2025",
    address = "Vienna, Austria",
    publisher = "Association for Computational Linguistics",
    url = "https://aclanthology.org/2025.acl-long.1029/",
    doi = "10.18653/v1/2025.acl-long.1029",
    pages = "21171--21188",
    ISBN = "979-8-89176-251-0"
}

@misc{ste,
      title={Estimating or Propagating Gradients Through Stochastic Neurons for Conditional Computation}, 
      author={Yoshua Bengio and Nicholas Léonard and Aaron Courville},
      year={2013},
      eprint={1308.3432},
      archivePrefix={arXiv},
      primaryClass={cs.LG},
      url={https://arxiv.org/abs/1308.3432}, 
}

@article{ou2025inference,
  title={Inference-Time Scaling of Discrete Diffusion Models via Importance Weighting and Optimal Proposal Design},
  author={Ou, Zijing and Pani, Chinmay and Li, Yingzhen},
  journal={arXiv e-prints},
  pages={arXiv--2505},
  year={2025}
}

@misc{d1,
      title={d1: Scaling Reasoning in Diffusion Large Language Models via Reinforcement Learning}, 
      author={Siyan Zhao and Devaansh Gupta and Qinqing Zheng and Aditya Grover},
      year={2025},
      eprint={2504.12216},
      archivePrefix={arXiv},
      primaryClass={cs.CL},
      url={https://arxiv.org/abs/2504.12216}, 
}

@misc{d2po,
      title={Preference-Based Alignment of Discrete Diffusion Models}, 
      author={Umberto Borso and Davide Paglieri and Jude Wells and Tim Rocktäschel},
      year={2025},
      eprint={2503.08295},
      archivePrefix={arXiv},
      primaryClass={cs.LG},
      url={https://arxiv.org/abs/2503.08295}, 
}

@misc{ddpp,
      title={Steering Masked Discrete Diffusion Models via Discrete Denoising Posterior Prediction}, 
      author={Jarrid Rector-Brooks and Mohsin Hasan and Zhangzhi Peng and Zachary Quinn and Chenghao Liu and Sarthak Mittal and Nouha Dziri and Michael Bronstein and Yoshua Bengio and Pranam Chatterjee and Alexander Tong and Avishek Joey Bose},
      year={2024},
      eprint={2410.08134},
      archivePrefix={arXiv},
      primaryClass={cs.LG},
      url={https://arxiv.org/abs/2410.08134}, 
}

@misc{gao2022scalinglawsrewardmodel,
      title={Scaling Laws for Reward Model Overoptimization}, 
      author={Leo Gao and John Schulman and Jacob Hilton},
      year={2022},
      eprint={2210.10760},
      archivePrefix={arXiv},
      primaryClass={cs.LG},
      url={https://arxiv.org/abs/2210.10760}, 
}

@misc{moskovitz2023confrontingrewardmodeloveroptimization,
      title={Confronting Reward Model Overoptimization with Constrained RLHF}, 
      author={Ted Moskovitz and Aaditya K. Singh and DJ Strouse and Tuomas Sandholm and Ruslan Salakhutdinov and Anca D. Dragan and Stephen McAleer},
      year={2023},
      eprint={2310.04373},
      archivePrefix={arXiv},
      primaryClass={cs.LG},
      url={https://arxiv.org/abs/2310.04373}, 
}

@misc{armorm,
      title={Interpretable Preferences via Multi-Objective Reward Modeling and Mixture-of-Experts}, 
      author={Haoxiang Wang and Wei Xiong and Tengyang Xie and Han Zhao and Tong Zhang},
      year={2024},
      eprint={2406.12845},
      archivePrefix={arXiv},
      primaryClass={cs.LG},
      url={https://arxiv.org/abs/2406.12845}, 
}

@misc{rlhf_openai,
      title={Training language models to follow instructions with human feedback}, 
      author={Long Ouyang and Jeff Wu and Xu Jiang and Diogo Almeida and Carroll L. Wainwright and Pamela Mishkin and Chong Zhang and Sandhini Agarwal and Katarina Slama and Alex Ray and John Schulman and Jacob Hilton and Fraser Kelton and Luke Miller and Maddie Simens and Amanda Askell and Peter Welinder and Paul Christiano and Jan Leike and Ryan Lowe},
      year={2022},
      eprint={2203.02155},
      archivePrefix={arXiv},
      primaryClass={cs.CL},
      url={https://arxiv.org/abs/2203.02155}, 
}

@misc{tr2d2,
      title={TR2-D2: Tree Search Guided Trajectory-Aware Fine-Tuning for Discrete Diffusion}, 
      author={Sophia Tang and Yuchen Zhu and Molei Tao and Pranam Chatterjee},
      year={2025},
      eprint={2509.25171},
      archivePrefix={arXiv},
      primaryClass={cs.LG},
      url={https://arxiv.org/abs/2509.25171}, 
}

@misc{fk-steering,
      title={A General Framework for Inference-time Scaling and Steering of Diffusion Models}, 
      author={Raghav Singhal and Zachary Horvitz and Ryan Teehan and Mengye Ren and Zhou Yu and Kathleen McKeown and Rajesh Ranganath},
      year={2025},
      eprint={2501.06848},
      archivePrefix={arXiv},
      primaryClass={cs.LG},
      url={https://arxiv.org/abs/2501.06848}, 
}

@misc{dts,
      title={Diffusion Tree Sampling: Scalable inference-time alignment of diffusion models}, 
      author={Vineet Jain and Kusha Sareen and Mohammad Pedramfar and Siamak Ravanbakhsh},
      year={2025},
      eprint={2506.20701},
      archivePrefix={arXiv},
      primaryClass={cs.LG},
      url={https://arxiv.org/abs/2506.20701}, 
}

@misc{dm_noise_trajectory,
      title={Test-Time Scaling of Diffusion Models via Noise Trajectory Search}, 
      author={Vignav Ramesh and Morteza Mardani},
      year={2025},
      eprint={2506.03164},
      archivePrefix={arXiv},
      primaryClass={cs.LG},
      url={https://arxiv.org/abs/2506.03164}, 
}

@misc{guo2025trainingfreeguidancedifferentiabilityscalable,
      title={Training-Free Guidance Beyond Differentiability: Scalable Path Steering with Tree Search in Diffusion and Flow Models}, 
      author={Yingqing Guo and Yukang Yang and Hui Yuan and Mengdi Wang},
      year={2025},
      eprint={2502.11420},
      archivePrefix={arXiv},
      primaryClass={cs.LG},
      url={https://arxiv.org/abs/2502.11420}, 
}

@misc{kim2025testtimealignmentdiffusionmodels,
      title={Test-time Alignment of Diffusion Models without Reward Over-optimization}, 
      author={Sunwoo Kim and Minkyu Kim and Dongmin Park},
      year={2025},
      eprint={2501.05803},
      archivePrefix={arXiv},
      primaryClass={cs.LG},
      url={https://arxiv.org/abs/2501.05803}, 
}

@misc{bansal2023universalguidancediffusionmodels,
      title={Universal Guidance for Diffusion Models}, 
      author={Arpit Bansal and Hong-Min Chu and Avi Schwarzschild and Soumyadip Sengupta and Micah Goldblum and Jonas Geiping and Tom Goldstein},
      year={2023},
      eprint={2302.07121},
      archivePrefix={arXiv},
      primaryClass={cs.CV},
      url={https://arxiv.org/abs/2302.07121}, 
}

@misc{zhang2025inferencetimescalingdiffusionmodels,
      title={Inference-time Scaling of Diffusion Models through Classical Search}, 
      author={Xiangcheng Zhang and Haowei Lin and Haotian Ye and James Zou and Jianzhu Ma and Yitao Liang and Yilun Du},
      year={2025},
      eprint={2505.23614},
      archivePrefix={arXiv},
      primaryClass={cs.LG},
      url={https://arxiv.org/abs/2505.23614}, 
}

@misc{zekri2025finetuningdiscretediffusionmodels,
      title={Fine-Tuning Discrete Diffusion Models with Policy Gradient Methods}, 
      author={Oussama Zekri and Nicolas Boullé},
      year={2025},
      eprint={2502.01384},
      archivePrefix={arXiv},
      primaryClass={stat.ML},
      url={https://arxiv.org/abs/2502.01384}, 
}

@misc{wang2025finetuningdiscretediffusionmodels,
      title={Fine-Tuning Discrete Diffusion Models via Reward Optimization with Applications to DNA and Protein Design}, 
      author={Chenyu Wang and Masatoshi Uehara and Yichun He and Amy Wang and Tommaso Biancalani and Avantika Lal and Tommi Jaakkola and Sergey Levine and Hanchen Wang and Aviv Regev},
      year={2025},
      eprint={2410.13643},
      archivePrefix={arXiv},
      primaryClass={cs.LG},
      url={https://arxiv.org/abs/2410.13643}, 
}

@misc{xie2025dreamcoder7bopendiffusion,
      title={Dream-Coder 7B: An Open Diffusion Language Model for Code}, 
      author={Zhihui Xie and Jiacheng Ye and Lin Zheng and Jiahui Gao and Jingwei Dong and Zirui Wu and Xueliang Zhao and Shansan Gong and Xin Jiang and Zhenguo Li and Lingpeng Kong},
      year={2025},
      eprint={2509.01142},
      archivePrefix={arXiv},
      primaryClass={cs.CL},
      url={https://arxiv.org/abs/2509.01142}, 
}

@misc{simple-dLLM,
      title={dLLM: Simple Diffusion Language Modeling}, 
      author={Zhanhui Zhou and Lingjie Chen and Hanghang Tong and Dawn Song},
      year={2026},
      eprint={2602.22661},
      archivePrefix={arXiv},
      primaryClass={cs.CL},
      url={https://arxiv.org/abs/2602.22661}, 
}

@misc{bhaskar2025languagemodelsthinkchat,
      title={Language Models that Think, Chat Better}, 
      author={Adithya Bhaskar and Xi Ye and Danqi Chen},
      year={2025},
      eprint={2509.20357},
      archivePrefix={arXiv},
      primaryClass={cs.CL},
      url={https://arxiv.org/abs/2509.20357}, 
}

@misc{zhao2024wildchat1mchatgptinteraction,
      title={WildChat: 1M ChatGPT Interaction Logs in the Wild}, 
      author={Wenting Zhao and Xiang Ren and Jack Hessel and Claire Cardie and Yejin Choi and Yuntian Deng},
      year={2024},
      eprint={2405.01470},
      archivePrefix={arXiv},
      primaryClass={cs.CL},
      url={https://arxiv.org/abs/2405.01470}, 
}

@misc{shao2024deepseekmathpushinglimitsmathematical,
      title={DeepSeekMath: Pushing the Limits of Mathematical Reasoning in Open Language Models}, 
      author={Zhihong Shao and Peiyi Wang and Qihao Zhu and Runxin Xu and Junxiao Song and Xiao Bi and Haowei Zhang and Mingchuan Zhang and Y. K. Li and Y. Wu and Daya Guo},
      year={2024},
      eprint={2402.03300},
      archivePrefix={arXiv},
      primaryClass={cs.CL},
      url={https://arxiv.org/abs/2402.03300}, 
}

@misc{prabhudesai2024videodiffusionalignmentreward,
      title={Video Diffusion Alignment via Reward Gradients}, 
      author={Mihir Prabhudesai and Russell Mendonca and Zheyang Qin and Katerina Fragkiadaki and Deepak Pathak},
      year={2024},
      eprint={2407.08737},
      archivePrefix={arXiv},
      primaryClass={cs.CV},
      url={https://arxiv.org/abs/2407.08737}, 
}

@misc{freedom,
      title={FreeDoM: Training-Free Energy-Guided Conditional Diffusion Model}, 
      author={Jiwen Yu and Yinhuai Wang and Chen Zhao and Bernard Ghanem and Jian Zhang},
      year={2023},
      eprint={2303.09833},
      archivePrefix={arXiv},
      primaryClass={cs.CV},
      url={https://arxiv.org/abs/2303.09833}, 
}

@misc{mpgd,
      title={Manifold Preserving Guided Diffusion}, 
      author={Yutong He and Naoki Murata and Chieh-Hsin Lai and Yuhta Takida and Toshimitsu Uesaka and Dongjun Kim and Wei-Hsiang Liao and Yuki Mitsufuji and J. Zico Kolter and Ruslan Salakhutdinov and Stefano Ermon},
      year={2023},
      eprint={2311.16424},
      archivePrefix={arXiv},
      primaryClass={cs.LG},
      url={https://arxiv.org/abs/2311.16424}, 
}

@misc{tfg,
      title={TFG: Unified Training-Free Guidance for Diffusion Models}, 
      author={Haotian Ye and Haowei Lin and Jiaqi Han and Minkai Xu and Sheng Liu and Yitao Liang and Jianzhu Ma and James Zou and Stefano Ermon},
      year={2024},
      eprint={2409.15761},
      archivePrefix={arXiv},
      primaryClass={cs.LG},
      url={https://arxiv.org/abs/2409.15761}, 
}

@misc{black2024trainingdiffusionmodelsreinforcement,
      title={Training Diffusion Models with Reinforcement Learning}, 
      author={Kevin Black and Michael Janner and Yilun Du and Ilya Kostrikov and Sergey Levine},
      year={2024},
      eprint={2305.13301},
      archivePrefix={arXiv},
      primaryClass={cs.LG},
      url={https://arxiv.org/abs/2305.13301}, 
}

@misc{fan2023dpokreinforcementlearningfinetuning,
      title={DPOK: Reinforcement Learning for Fine-tuning Text-to-Image Diffusion Models}, 
      author={Ying Fan and Olivia Watkins and Yuqing Du and Hao Liu and Moonkyung Ryu and Craig Boutilier and Pieter Abbeel and Mohammad Ghavamzadeh and Kangwook Lee and Kimin Lee},
      year={2023},
      eprint={2305.16381},
      archivePrefix={arXiv},
      primaryClass={cs.LG},
      url={https://arxiv.org/abs/2305.16381}, 
}

@misc{clark2024directlyfinetuningdiffusionmodels,
      title={Directly Fine-Tuning Diffusion Models on Differentiable Rewards}, 
      author={Kevin Clark and Paul Vicol and Kevin Swersky and David J Fleet},
      year={2024},
      eprint={2309.17400},
      archivePrefix={arXiv},
      primaryClass={cs.CV},
      url={https://arxiv.org/abs/2309.17400}, 
}

@misc{xu2023imagerewardlearningevaluatinghuman,
      title={ImageReward: Learning and Evaluating Human Preferences for Text-to-Image Generation}, 
      author={Jiazheng Xu and Xiao Liu and Yuchen Wu and Yuxuan Tong and Qinkai Li and Ming Ding and Jie Tang and Yuxiao Dong},
      year={2023},
      eprint={2304.05977},
      archivePrefix={arXiv},
      primaryClass={cs.CV},
      url={https://arxiv.org/abs/2304.05977}, 
}

@misc{anil2026finetuningdiffusionmodelsintermediate,
      title={Fine-Tuning Diffusion Models via Intermediate Distribution Shaping}, 
      author={Gautham Govind Anil and Shaan Ul Haque and Nithish Kannen and Dheeraj Nagaraj and Sanjay Shakkottai and Karthikeyan Shanmugam},
      year={2026},
      eprint={2510.02692},
      archivePrefix={arXiv},
      primaryClass={cs.LG},
      url={https://arxiv.org/abs/2510.02692}, 
}

@misc{zheng2024lmsyschat1mlargescalerealworldllm,
      title={LMSYS-Chat-1M: A Large-Scale Real-World LLM Conversation Dataset}, 
      author={Lianmin Zheng and Wei-Lin Chiang and Ying Sheng and Tianle Li and Siyuan Zhuang and Zhanghao Wu and Yonghao Zhuang and Zhuohan Li and Zi Lin and Eric P. Xing and Joseph E. Gonzalez and Ion Stoica and Hao Zhang},
      year={2024},
      eprint={2309.11998},
      archivePrefix={arXiv},
      primaryClass={cs.CL},
      url={https://arxiv.org/abs/2309.11998}, 
}

@misc{xu2024magpiealignmentdatasynthesis,
      title={Magpie: Alignment Data Synthesis from Scratch by Prompting Aligned LLMs with Nothing}, 
      author={Zhangchen Xu and Fengqing Jiang and Luyao Niu and Yuntian Deng and Radha Poovendran and Yejin Choi and Bill Yuchen Lin},
      year={2024},
      eprint={2406.08464},
      archivePrefix={arXiv},
      primaryClass={cs.CL},
      url={https://arxiv.org/abs/2406.08464}, 
}


\appendix
\newpage
\section{Appendix}
The appendix is organized as follows: In \autoref{sec:appdx_exp_setup}, we present implementation details such as prompts, hyperparameters, and compute. In \autoref{sec:appdx_more_results}, we present additional results and raw values used to generate plots and figures.\\

\section{Experimental Setup}
\label{sec:appdx_exp_setup}

\subsection{Implementation Details.} 
\textbf{Test-Time Adaptation.} We perform all experiments on 4 H100 GPUs. We report averaged results over 5 seeds (and standard errors) comprising a subset of 320 prompts per dataset. We generate sequences up to length 128 tokens, decoding 1 token for each denoising step. We set $\eta$=0.5, $M$=3, and $N$=4. Unless stated otherwise, $\tau=0.7$. For all methods, we deprioritize the \texttt{EOS} token to the lowest priority,  similar to  \citet{xie2025dreamcoder7bopendiffusion}, as we noticed that it leads to improved performance even for the BoN baseline.

\textbf{Training Experiments.} We perform all experiments on 2 GH200 GPUs. We report results averaged over 3 independent runs (with standard error bands). We adapt the codebase from \citet{simple-dLLM}\footnote{\url{https://github.com/ZHZisZZ/dllm}}, which provides an implementation of the diffu-GRPO RL algorithm \citep{d1}. We train for 500 steps with a batch size of 4. We generate $N$=4 completions per prompt of length 128, decoding 1 token at a time, with a sampling temperature of 0.9 following \citep{d1}. For diffu-GRPO \citep{d1}, we set KL-$\beta=0.04$ and clipping ratio $\epsilon=0.2$. For RGRL, we use $M=1$, $\eta=0.5$. For all methods, we train with learning rate $5e-6$, and use LoRA with $r=32$, $\alpha=32$, dropout $0.1$.

\textbf{LMUnit evaluation.} We evaluate response quality using LMUnit \citep{lmunit}, specifically the \textit{LMUnit-Qwen2.5-72B} model served via the official \texttt{lmunit} library at \url{https://github.com/ContextualAI/LMUnit}. Following the official inference protocol, we use greedy decoding with \texttt{logprobs=20} to obtain continuous scores on a 1--5 scale. Each response is evaluated against five unit tests covering relevance, correctness, coherence, and safety, as elaborated in~\appref{subsec:appdx_model_inputs}. The final  score is computed as the average across all unit tests.

\textbf{Reward Models.} Our experiments encompass the following reward models: \textit{Skywork-Reward-V2-Qwen3-0.6B}\footnote{\href{https://huggingface.co/Skywork/Skywork-Reward-V2-Qwen3-0.6B}{Skywork/Skywork-Reward-V2-Qwen3-0.6B}}, \textit{Skywork-Reward-V2-Qwen3-1.7B}\footnote{\href{https://huggingface.co/Skywork/Skywork-Reward-V2-Qwen3-1.7B}{Skywork/Skywork-Reward-V2-Qwen3-1.7B}}, \textit{Skywork-Reward-V2-Qwen3-4B}\footnote{\href{https://huggingface.co/Skywork/Skywork-Reward-V2-Qwen3-4B}{Skywork/Skywork-Reward-V2-Qwen3-4B}}, and \textit{Skywork-Reward-V2-Llama-3.2-1B}.

\textbf{Test-Time Adaptation Datasets.} For the test-time adaptation experiments, we use Reward-Bench-2~\citep{rewardbench2}\footnote{\url{https://huggingface.co/datasets/allenai/reward-bench-2}}, RM-Bench~\citep{rmbench}\footnote{\url{https://huggingface.co/datasets/THU-KEG/RM-Bench}}, and JudgeBench~\citep{judgebench}\footnote{\url{https://huggingface.co/datasets/ScalerLab/JudgeBench}}.

\textbf{Post-Training Datasets.} We use WildChat-IF~\citep{bhaskar2025languagemodelsthinkchat, zhao2024wildchat1mchatgptinteraction}\footnote{\url{allenai/tulu-3-wildchat-if-on-policy-8b}}, and lmsys-chat-1m~\citep{zheng2024lmsyschat1mlargescalerealworldllm}\footnote{\url{https://huggingface.co/datasets/lmsys/lmsys-chat-1m}}. For lmsys-chat-1m we use the train split, and filter by the English subset. We filter Magpie-Ultra \citep{xu2024magpiealignmentdatasynthesis}\footnote{\url{https://huggingface.co/datasets/argilla/magpie-ultra-v0.1}} by quality ``good'' and above from the train set. 

\subsection{Model Inputs}
\label{subsec:appdx_model_inputs}

\autoref{fig:templates_reward_and_diffusion} shows the prompt templates used for \textit{Dream-v0-Instruct-7B} \citep{dream} and the \textit{Skywork-Reward-v2} \citep{skywork} reward models. \autoref{fig:templates_lmunit} shows the prompt template and unit tests used for LMUnit \citep{lmunit}.

\begin{figure}[h]
\begin{tcolorbox}[
  title=Input Templates,
  colback=gray!5,
  colframe=gray!100,
  fonttitle=\bfseries,
  boxrule=0.5pt,
]
\textbf{Diffusion Model (Generation):}
\begin{tcolorbox}[colback=white, colframe=gray!30, boxrule=0.3pt, left=4pt, right=4pt, top=2pt, bottom=2pt]
\ttfamily\small
<|im\_start|>user\\
\{prompt\}<|im\_end|>\\
<|im\_start|>assistant\\
\{generated response\}
\end{tcolorbox}
\vspace{6pt}
\textbf{Reward Model -- Soft Scoring (During Optimization):}
\begin{tcolorbox}[colback=white, colframe=gray!30, boxrule=0.3pt, left=4pt, right=4pt, top=2pt, bottom=2pt]
\ttfamily\small
<|im\_start|>user\\
\{prompt\}<|im\_end|>\\
<|im\_start|>assistant\\
\{response embeddings\}<|im\_end|>
\end{tcolorbox}
\vspace{6pt}
\textbf{Reward Model -- Discrete Scoring:}
\begin{tcolorbox}[colback=white, colframe=gray!30, boxrule=0.3pt, left=4pt, right=4pt, top=2pt, bottom=2pt]
\ttfamily\small
<|im\_start|>user\\
\{prompt\}<|im\_end|>\\
<|im\_start|>assistant\\
\{response\}<|im\_end|>
\end{tcolorbox}
\end{tcolorbox}
\caption{Input templates for the diffusion model and reward model.}
\label{fig:templates_reward_and_diffusion}
\end{figure}

\begin{figure}[h]
\begin{tcolorbox}[
  title=LMUnit Evaluation Prompts,
  colback=gray!5,
  colframe=gray!100,
  fonttitle=\bfseries,
  boxrule=0.5pt,
]
Each response is evaluated using 5 prompts of the following form:

\vspace{6pt}
\begin{tcolorbox}[
  colback=white,
  colframe=gray!30,
  boxrule=0.4pt,
  left=4pt, right=4pt, top=4pt, bottom=4pt,
]
\ttfamily\small
Query: \{query\}\\
Response: \{response\}\\
Unit Test: \{unit\_test\}
\end{tcolorbox}
\vspace{4pt}

\noindent where \texttt{\{unit\_test\}} is one of:
\begin{enumerate}[noitemsep, topsep=4pt, leftmargin=2em, label=(\arabic*)]
    \item Does the response directly and effectively address the user's request?
    \item Is the information in the response correct and reliable?
    \item Is the response well-structured, clear, and fluent?
    \item Does the response appropriately address the full scope of the question?
    \item Is the response free from harmful, biased, or inappropriate content?
\end{enumerate}
\end{tcolorbox}
\caption{Input template and unit tests for LMUnit.}
\label{fig:templates_lmunit}
\end{figure}

\section{Additional Results}
\label{sec:appdx_more_results}

\begin{table*}[h]
\centering
\footnotesize
\addtolength{\tabcolsep}{-2.5pt}
\caption{Performance of \textit{Dream-v0-7B-Instruct} on Reward-Bench-2 \citep{rewardbench2}, JudgeBench \citep{judgebench}, and RM-Bench \citep{rmbench} with varying reward model sizes $(\tau=0.7)$.}
\label{tab:rm_size_ablation}
\begin{tabular}{lccccccccc}
\toprule
\multirow{2}{*}{\textbf{Method}} & \multicolumn{3}{c}{Reward-Bench-2} & \multicolumn{3}{c}{JudgeBench} & \multicolumn{3}{c}{RM-Bench} \\ \cmidrule(lr){2-4} \cmidrule(lr){5-7} \cmidrule(lr){8-10}
 & \textbf{Top@1} & \textbf{Avg@4} & \textbf{LMUnit} & \textbf{Top@1} & \textbf{Avg@4} & \textbf{LMUnit} & \textbf{Top@1} & \textbf{Avg@4} & \textbf{LMUnit} \\ \midrule
\multicolumn{10}{c}{\textit{Skywork-Reward-v2-Qwen3-0.6B}} \\ \midrule
BoN & 2.29\err{0.16} & 0.96\err{0.19} & 4.13\err{0.01} & 2.01\err{0.13} & -0.24\err{0.16} & 3.88\err{0.02} & 4.09\err{0.18} & 2.32\err{0.16} & \textbf{4.01}\err{0.04} \\
Expectation & {\ul 2.71}\err{0.26} & {\ul 1.49}\err{0.25} & \textbf{4.18}\err{0.03} & \textbf{2.56}\err{0.06} & \textbf{0.49}\err{0.08} & \textbf{3.91}\err{0.03} & \textbf{4.49}\err{0.14} & \textbf{2.67}\err{0.12} & \textbf{4.00}\err{0.01} \\
APS & 2.64\err{0.21} & 1.22\err{0.20} & \textbf{4.18}\err{0.03} & 2.21\err{0.06} & -0.02\err{0.11} & 3.86\err{0.01} & 4.21\err{0.14} & 2.28\err{0.10} & 3.95\err{0.02} \\
\rowcolor{highlight} EntRGi & \textbf{3.07}\err{0.18} & \textbf{1.65}\err{0.17} & \textbf{4.21}\err{0.02} & \textbf{2.50}\err{0.10} & \textbf{0.41}\err{0.10} & \textbf{3.92}\err{0.02} & \textbf{4.49}\err{0.06} & {\ul 2.54}\err{0.12} & 3.98\err{0.01} \\ \midrule
\multicolumn{10}{c}{\textit{Skywork-Reward-v2-Qwen3-4B}} \\ \midrule
BoN & 10.27\err{0.39} & 7.99\err{0.39} & 4.15\err{0.01} & 7.68\err{0.07} & 4.76\err{0.14} & 3.92\err{0.02} & 13.03\err{0.28} & 10.72\err{0.24} & 4.06\err{0.03} \\
Expectation & \textbf{11.35}\err{0.34} & \textbf{9.23}\err{0.31} & \textbf{4.28}\err{0.03} & {\ul 8.39}\err{0.18} & \textbf{5.69}\err{0.16} & {\ul 3.93}\err{0.01} & {\ul 13.39}\err{0.21} & \textbf{10.96}\err{0.24} & {\ul 4.07}\err{0.02} \\
APS & 11.11\err{0.36} & 8.80\err{0.35} & 4.26\err{0.02} & 8.12\err{0.18} & 5.13\err{0.09} & {\ul 3.93}\err{0.02} & 13.11\err{0.23} & 10.48\err{0.18} & 4.05\err{0.02} \\
\rowcolor{highlight} EntRGi & \textbf{11.40}\err{0.27} & \textbf{9.26}\err{0.35} & \textbf{4.29}\err{0.01} & \textbf{8.60}\err{0.12} & \textbf{5.78}\err{0.10} & \textbf{3.97}\err{0.03} & \textbf{13.67}\err{0.15} & \textbf{11.10}\err{0.22} & \textbf{4.09}\err{0.02} \\ \bottomrule
\end{tabular}
\end{table*}
\subsection{Scaling Reward Model Size}
\label{sec:appdx_rm_size}
\autoref{tab:rm_size_ablation} presents results on two additional reward models, \textit{Skywork-Reward-v2-0.6B} and \textit{Skywork-Reward-v2-4B}. Results with \textit{Skywork-Reward-v2-1.7B} are presented in \autoref{tab:main_results} in the main paper. We observe similar trends for all 3 models, as shown in ~\autoref{fig:scaling_rm_size} in the main paper.

\begin{table*}[h]
\centering
\footnotesize
\addtolength{\tabcolsep}{-1.3pt}
\caption{Effect of gradient steps $M$ on performance with \textit{Skywork-Reward-v2-Qwen3-0.6B} ($\tau=0.7$).}
\label{tab:m_scaling_0.6b}
\begin{tabular}{lccccccccc}
\toprule
\multirow{2}{*}{\textbf{Method}} & \multicolumn{3}{c}{Reward-Bench-2} & \multicolumn{3}{c}{JudgeBench} & \multicolumn{3}{c}{RM-Bench} \\ \cmidrule(lr){2-4} \cmidrule(lr){5-7} \cmidrule(lr){8-10}
 & \textbf{Top@1} & \textbf{Avg@4} & \textbf{LMUnit} & \textbf{Top@1} & \textbf{Avg@4} & \textbf{LMUnit} & \textbf{Top@1} & \textbf{Avg@4} & \textbf{LMUnit} \\ \midrule
$M=1$ & 2.82\err{0.24} & 1.36\err{0.31} & 4.28\err{0.02} & 2.20\err{0.30} & 0.14\err{0.29} & 3.95\err{0.05} & 4.37\err{0.21} & 2.50\err{0.18} & 4.05\err{0.02} \\
$M=2$ & 2.97\err{0.32} & 1.62\err{0.33} & 4.20\err{0.02} & 2.22\err{0.32} & 0.23\err{0.25} & 3.95\err{0.09} & 4.37\err{0.16} & 2.33\err{0.11} & 4.08\err{0.04} \\
$M=3$ & 3.17\err{0.24} & 1.89\err{0.29} & 4.27\err{0.05} & 2.82\err{0.20} & 0.61\err{0.21} & 4.00\err{0.05} & 4.31\err{0.22} & 2.23\err{0.19} & 4.00\err{0.04} \\
$M=4$ & 3.30\err{0.28} & 1.91\err{0.35} & 4.26\err{0.04} & 2.83\err{0.26} & 0.40\err{0.26} & 3.93\err{0.06} & 4.62\err{0.08} & 2.62\err{0.15} & 4.07\err{0.03} \\
$M=5$ & 3.25\err{0.37} & 1.95\err{0.40} & 4.30\err{0.03} & 2.69\err{0.22} & 0.53\err{0.26} & 3.93\err{0.04} & 4.67\err{0.22} & 2.63\err{0.22} & 3.99\err{0.02} \\ \bottomrule
\end{tabular}
\end{table*}

\begin{table*}[h]
\centering
\footnotesize
\addtolength{\tabcolsep}{-1.5pt}
\caption{Effect of gradient steps $M$ on performance with \textit{Skywork-Reward-v2-Qwen3-1.7B} ($\tau=0.7$).}
\label{tab:m_scaling_1.7b}
\begin{tabular}{lccccccccc}
\toprule
\multirow{2}{*}{\textbf{Method}} & \multicolumn{3}{c}{Reward-Bench-2} & \multicolumn{3}{c}{JudgeBench} & \multicolumn{3}{c}{RM-Bench} \\ \cmidrule(lr){2-4} \cmidrule(lr){5-7} \cmidrule(lr){8-10}
 & \textbf{Top@1} & \textbf{Avg@4} & \textbf{LMUnit} & \textbf{Top@1} & \textbf{Avg@4} & \textbf{LMUnit} & \textbf{Top@1} & \textbf{Avg@4} & \textbf{LMUnit} \\ \midrule
$M=1$ & 3.74\err{0.44} & 2.06\err{0.52} & 4.27\err{0.05} & 2.34\err{0.13} & -0.13\err{0.19} & 3.99\err{0.05} & 5.05\err{0.35} & 2.94\err{0.31} & 4.09\err{0.02} \\
$M=2$ & 4.07\err{0.41} & 2.29\err{0.48} & 4.33\err{0.04} & 2.45\err{0.27} & -0.04\err{0.24} & 3.98\err{0.05} & 5.44\err{0.40} & 3.16\err{0.33} & 4.09\err{0.03} \\
$M=3$ & 4.13\err{0.48} & 2.65\err{0.46} & 4.25\err{0.03} & 2.72\err{0.15} & 0.22\err{0.23} & 4.03\err{0.05} & 5.86\err{0.38} & 3.44\err{0.34} & 4.12\err{0.04} \\
$M=4$ & 4.55\err{0.46} & 2.71\err{0.55} & 4.30\err{0.03} & 2.98\err{0.38} & 0.46\err{0.25} & 3.95\err{0.05} & 6.06\err{0.38} & 3.48\err{0.38} & 4.14\err{0.05} \\
$M=5$ & 4.90\err{0.63} & 2.94\err{0.56} & 4.29\err{0.05} & 2.70\err{0.32} & 0.46\err{0.26} & 4.00\err{0.03} & 5.71\err{0.34} & 3.03\err{0.35} & 4.05\err{0.05} \\ \bottomrule
\end{tabular}
\end{table*}

\begin{table*}[h]
\centering
\footnotesize
\addtolength{\tabcolsep}{-1.9pt}
\caption{Effect of gradient steps $M$ on performance with \textit{Skywork-Reward-v2-Qwen3-4B} ($\tau=0.7$).}
\label{tab:m_scaling_4b}
\begin{tabular}{lccccccccc}
\toprule
\multirow{2}{*}{\textbf{Method}} & \multicolumn{3}{c}{Reward-Bench-2} & \multicolumn{3}{c}{JudgeBench} & \multicolumn{3}{c}{RM-Bench} \\ \cmidrule(lr){2-4} \cmidrule(lr){5-7} \cmidrule(lr){8-10}
 & \textbf{Top@1} & \textbf{Avg@4} & \textbf{LMUnit} & \textbf{Top@1} & \textbf{Avg@4} & \textbf{LMUnit} & \textbf{Top@1} & \textbf{Avg@4} & \textbf{LMUnit} \\ \midrule
$M=1$ & 11.58\err{0.97} & 9.50\err{0.99} & 4.28\err{0.04} & 8.38\err{0.28} & 5.46\err{0.37} & 3.94\err{0.02} & 13.22\err{0.22} & 10.51\err{0.11} & 4.10\err{0.03} \\
$M=2$ & 11.61\err{0.81} & 9.55\err{0.82} & 4.29\err{0.04} & 8.66\err{0.36} & 5.83\err{0.31} & 3.96\err{0.06} & 13.23\err{0.06} & 10.92\err{0.18} & 4.15\err{0.03} \\
$M=3$ & 12.25\err{0.72} & 10.08\err{0.75} & 4.27\err{0.02} & 8.45\err{0.31} & 5.71\err{0.33} & 4.02\err{0.04} & 13.49\err{0.18} & 11.05\err{0.22} & 4.11\err{0.03} \\
$M=4$ & 12.13\err{0.71} & 10.00\err{0.78} & 4.33\err{0.05} & 8.64\err{0.26} & 6.03\err{0.29} & 4.00\err{0.05} & 13.47\err{0.23} & 11.11\err{0.15} & 4.16\err{0.03} \\
$M=5$ & 12.21\err{0.70} & 10.25\err{0.72} & 4.34\err{0.02} & 8.44\err{0.26} & 5.74\err{0.31} & 3.95\err{0.06} & 13.43\err{0.25} & 10.83\err{0.20} & 4.06\err{0.06} \\ \bottomrule
\end{tabular}
\end{table*}
\subsection{Scaling Reward Model Iterations}
\label{sec:appdx_m_scaling}
\autoref{tab:m_scaling_0.6b}, \autoref{tab:m_scaling_1.7b}, and \autoref{tab:m_scaling_4b} present results with scaling reward model guidance steps $M$ from 1 to 5 on all three reward models: \textit{Skywork-Reward-v2-0.6B}, \textit{Skywork-Reward-v2-1.7B}, and \textit{Skywork-Reward-v2-4B}. Aggregated results are presented in \autoref{fig:scaling_m} in the main paper. We observe similar trends across all reward models i.e. increasing $M$ increasing reward but is prone to reward hacking after a certain point. The optimal $M$ varies by dataset. All our main experiments are conducted using a fixed $M=3$ for all datasets.

\begin{table*}[h]
\centering
\footnotesize
\addtolength{\tabcolsep}{-2pt}
\caption{Performance of \textit{Dream-v0-7B-Instruct} with alternate weighting schemes on Reward-Bench-2 \citep{rewardbench2}, JudgeBench \citep{judgebench}, and RM-Bench \citep{rmbench} with varying reward model sizes $(\tau=0.7)$.}
\label{tab:weighting_appdx}
\begin{tabular}{lccccccccc}
\toprule
\multirow{2}{*}{\textbf{Method}} & \multicolumn{3}{c}{RewardBench-2} & \multicolumn{3}{c}{JudgeBench} & \multicolumn{3}{c}{RM-Bench} \\ \cmidrule(l){2-10} 
 & \textbf{Top@1} & \textbf{Avg@4} & \textbf{LMUnit} & \textbf{Top@1} & \textbf{Avg@4} & \textbf{LMUnit} & \textbf{Top@1} & \textbf{Avg@4} & \textbf{LMUnit} \\ \midrule
Expectation & \textbf{3.95}\err{0.28} & \textbf{2.23}\err{0.24} & {\ul 4.22}\err{0.02} & {\ul 2.30}\err{0.08} & \textbf{0.13}\err{0.07} & \textbf{3.97}\err{0.01} & 5.45\err{0.16} & {\ul 3.29}\err{0.13} & {\ul 4.02}\err{0.03} \\
APS & 3.62\err{0.27} & 1.80\err{0.24} & {\ul 4.22}\err{0.02} & 1.87\err{0.14} & -0.63\err{0.10} & 3.93\err{0.02} & 5.11\err{0.14} & 2.66\err{0.15} & 4.00\err{0.02} \\ \midrule
\rowcolor{highlight} Inv-EntRGi & 3.58\err{0.28} & 1.79\err{0.25} & {\ul 4.22}\err{0.02} & 1.84\err{0.15} & -0.59\err{0.14} & 3.90\err{0.03} & 5.24\err{0.15} & 2.82\err{0.21} & 4.00\err{0.01} \\
\rowcolor{highlight} L2-Norm & 3.72\err{0.23} & 1.99\err{0.21} & {\ul 4.22}\err{0.02} & 1.98\err{0.15} & -0.33\err{0.12} & 3.93\err{0.03} & \textbf{5.52}\err{0.17} & 3.09\err{0.20} & {\ul 4.02}\err{0.01} \\
\rowcolor{highlight} EntRGi & \textbf{3.91}\err{0.30} & \textbf{2.20}\err{0.26} & \textbf{4.25}\err{0.02} & \textbf{2.44}\err{0.06} & {\ul 0.02}\err{0.10} & \textbf{3.98}\err{0.02} & \textbf{5.70}\err{0.12} & \textbf{3.41}\err{0.14} & \textbf{4.04}\err{0.01} \\ \bottomrule
\end{tabular}
\end{table*}
\begin{figure*}[t]
  \centering
  \includegraphics[width=\linewidth]{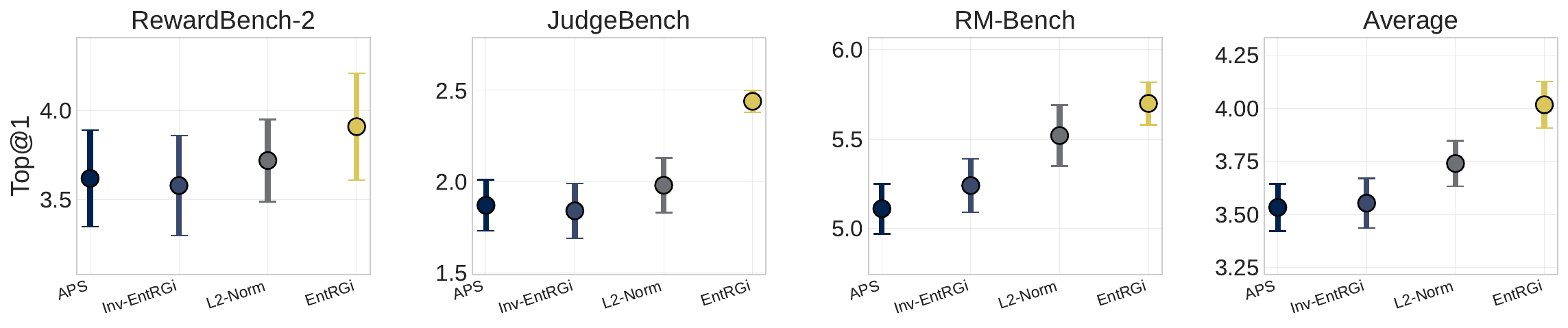}
  \caption{Comparison of token-level weighting mechanisms for EntRGi. We evaluate entropy-based weighting against inverse-entropy weighting Inv-EntGRi ($w^l = 1 - H(\vq^l) / \log K$) and an L2-norm heuristic ($w^l = \| {\ve}_\text{hard}^l - {\ve}_\text{soft}^l \| / \max_{l'} \| {\ve}_\text{hard}^{l'} - {\ve}_\text{soft}^{l'} \|$). Inverse-entropy weighting doesn't show noticeable improvements, while L2-norm-based weighting improves over APS but does not match regular EntRGi ($w^l = H(\vq^l) / \log K$). Raw scores are reported in \autoref{sec:appdx_weighting}.}
\label{fig:weighting_comparison}
\end{figure*}
\subsection{Weighting Mechanism}
\label{sec:appdx_weighting}

A natural question is whether EntRGi’s entropy-based weighting can be replaced by alternative signals, such as the L2 approximation error itself. \autoref{fig:weighting_comparison} and \autoref{tab:weighting_appdx} compare several weighting mechanisms. In Inv-EntRGi, higher entropy increases reliance on the soft relaxation, while in the L2-norm variant, token weights $w^l$ are derived from the L2 distance between hard and soft embeddings, normalized by the highest L2 norm at the sequence level.
We find that Inv-EntRGi consistently underperforms, and the L2-norm approach, while better than APS, does not match EntRGi. We believe that this is because normalized token entropy provides a naturally comparable signal across tokens and sequences, while L2 distances are unbounded and may require careful tuning.

\begin{table*}[h]
\centering
\footnotesize
\addtolength{\tabcolsep}{-1.5pt}
\caption{Performance of \textit{Dream-v0-7B-Instruct} on Reward-Bench-2 \citep{rewardbench2}, JudgeBench \citep{judgebench}, and RM-Bench \citep{rmbench} after decreasining denoising steps to 64 from 128.}
\label{tab:ablation_T}
\begin{tabular}{lccccccccc}
\toprule
\multirow{2}{*}{\textbf{Method}} & \multicolumn{3}{c}{RewardBench-2} & \multicolumn{3}{c}{JudgeBench} & \multicolumn{3}{c}{RM-Bench} \\ \cmidrule(lr){2-4} \cmidrule(lr){5-7} \cmidrule(lr){8-10}
 & \textbf{Top@1} & \textbf{Avg@4} & \textbf{LMUnit} & \textbf{Top@1} & \textbf{Avg@4} & \textbf{LMUnit} & \textbf{Top@1} & \textbf{Avg@4} & \textbf{LMUnit} \\ 
\midrule
\multicolumn{10}{c}{\textit{T=64}} \\ 
\midrule
BoN & 1.30\err{0.29} & -0.90\err{0.27} & 3.80\err{0.02} & 0.29\err{0.08} & -2.52\err{0.11} & 3.68\err{0.03} & 3.44\err{0.16} & 0.45\err{0.18} & 3.74\err{0.04} \\
\rowcolor{highlight} EntRGi & \textbf{2.34}\err{0.21} & \textbf{0.15}\err{0.22} & \textbf{3.96}\err{0.04} & \textbf{0.80}\err{0.11} & \textbf{-1.94}\err{0.13} & \textbf{3.70}\err{0.03} & \textbf{3.56}\err{0.25} & \textbf{0.63}\err{0.22} & 3.72\err{0.02} \\
\midrule
\multicolumn{10}{c}{\textit{T=128}} \\ \midrule
BoN & 2.99\err{0.23} & 1.38\err{0.29} & 4.15\err{0.02} & 1.65\err{0.18} & -0.84\err{0.16} & 3.91\err{0.02} & 5.11\err{0.20} & 2.98\err{0.15} & 4.02\err{0.03} \\
\rowcolor{highlight} EntRGi & \textbf{3.91}\err{0.30} & \textbf{2.20}\err{0.26} & \textbf{4.25}\err{0.02} & \textbf{2.44}\err{0.06} & {\ul 0.02}\err{0.10} & \textbf{3.98}\err{0.02} & \textbf{5.70}\err{0.12} & \textbf{3.41}\err{0.14} & \textbf{4.04}\err{0.01} \\
\bottomrule
\end{tabular}
\end{table*}
\subsection{Timestep Ablation}
\autoref{tab:ablation_T} reports results obtained by reducing the number of denoising timesteps from 128 to 64. The results show that the benefits of EntRGi’s gradient guidance persist even at lower denoising steps. For best performance, we recommend applying EntRGi at the highest number of denoising timesteps available.

\begin{figure*}[t]
  \centering
  \includegraphics[width=0.95\linewidth]{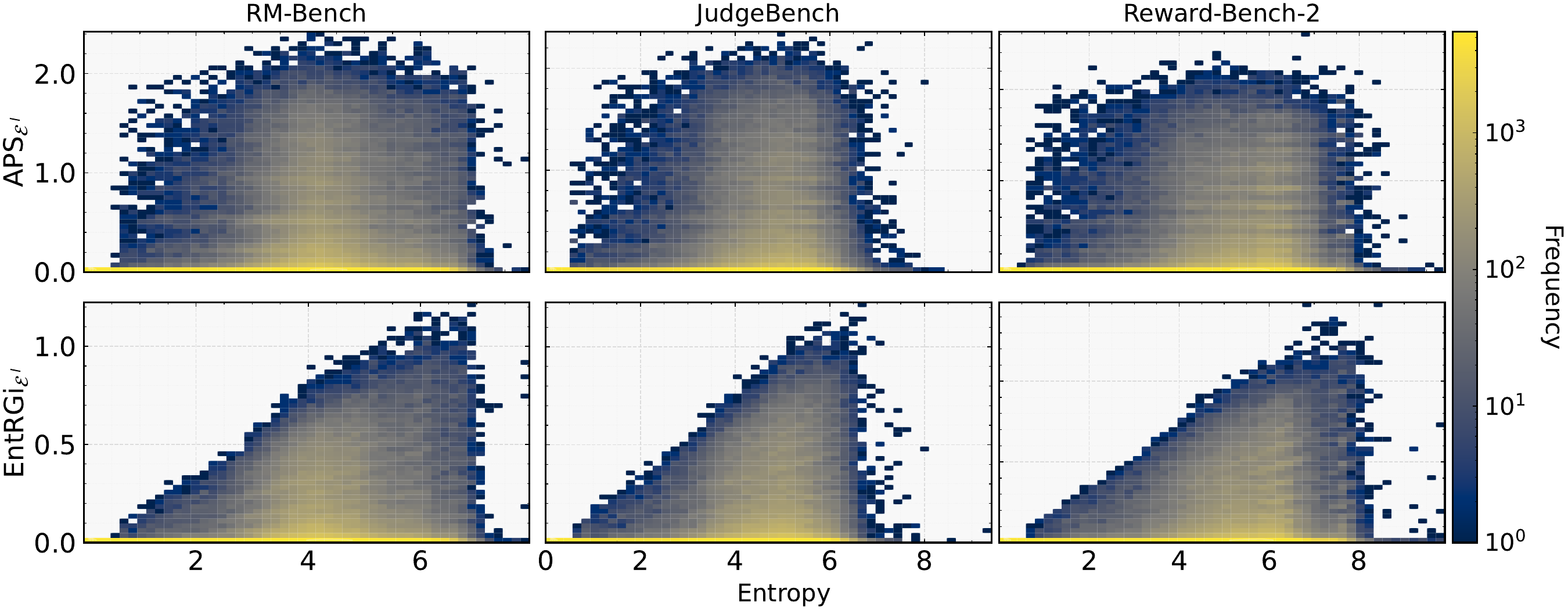}
  \caption{Heatmaps showing the joint distribution of entropy and approximation error $\mathcal{E}^l$ for three benchmarks (RM-Bench, JudgeBench, Reward-Bench-2) using APS (top) and EntRGi (bottom). Color indicates frequency on a log scale. EntRGi upweights soft tokens based on entropy. For entropy in the range 1--4, the soft approximation $\ve_\text{soft}$ is heavily preferred, trading off $\gE^l$ for $\gD^l$ proportionally.}
  \label{fig:error_aps_vs_eguide}
\end{figure*}
\subsection{EntRGi Error Analysis}
\label{subsec:appdx_error_histogram}
To understand the source of EntRGi’s gains, we analyze the relationship between predictive entropy and approximation error. \autoref{fig:error_aps_vs_eguide} visualizes the joint distribution of entropy and approximation error across three datasets. For APS (top row), approximation error grows sharply with entropy, indicating a strong mismatch between the discretized reward inputs and the continuous logits being updated. This steep error–entropy coupling leads to unreliable gradient signals.

In contrast, EntRGi (bottom row) exhibits a controlled and approximately linear error–entropy relationship. By adaptively reweighting soft embeddings and hard tokens at the token level, EntRGi limits approximation error in moderate-entropy regions while preserving reward-model fidelity at high entropy. This entropy-aware balancing produces more stable and reliable reward gradients, which directly translates into improved generation performance.

\subsection{Handling Tokenizer Mismatch}
In \autoref{sec:method}, we describe a simple approach to handle tokenizer mismatch by setting the embeddings of all non-overlapping tokens to zero. This situation arises in \textit{LLaDA-8B-Instruct}, since LLaDA is trained with a custom tokenizer. In contrast, most existing reward models are built on autoregressive (AR) backbones adapted for classification, and therefore do not share the same tokenizer. In particular, 45\%-55\% of LLaDA's tokenizer is mismatched with that of Qwen3/Llama-3. This mismatch does not occur for \textit{Dream-v0-7B-Instruct} as it uses a Qwen2.5 backbone \citep{dream}.
As discrete diffusion models become more widely adopted, we expect to see more reward models trained using dLLMs like LLaDA as the initialization, mitigating this discrepancy. Nevertheless, we present results using this formulation in ~\autoref{tab:tokenizer_results_llada_appdx}. Despite the mismatch, gradient-based methods remain effective and outperform BoN, which avoids this issue by decoding to text and re-encoding. Among all methods, EntRGi and RGRL-EntRGi achieve the best performance.

\begin{figure*}[h]
  \centering
  \includegraphics[width=\linewidth]{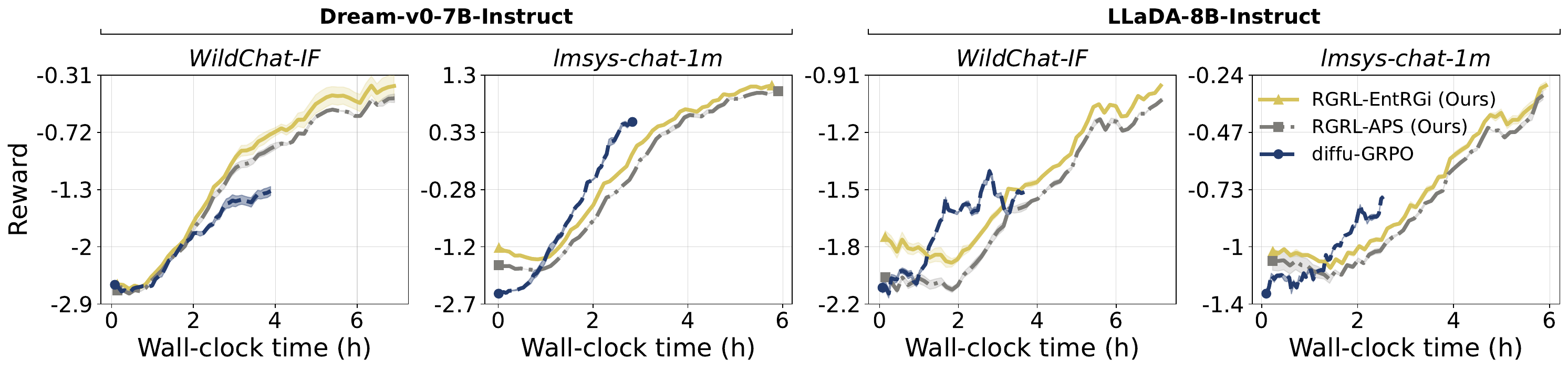}
  \caption{Reward vs. Wall-clock time curves when training on WildChat-IF and  \citep{bhaskar2025languagemodelsthinkchat, zhao2024wildchat1mchatgptinteraction} and lmsys-chat-1m \citep{zheng2024lmsyschat1mlargescalerealworldllm} with reward as per \textit{Skywork-Reward-V2-Qwen3-0.6B}.}
  \label{fig:rl_experiment_appdx}
\end{figure*}
\begin{table*}[h]
\centering
\footnotesize
\addtolength{\tabcolsep}{-1.2pt}
\caption{\textit{Dream-7B-Instruct} throughput on 32 randomly selected prompts from Reward-Bench-2, averaged over 3 seeds. All $N$ generations are decoded in parallel.}
\label{tab:throughput_comparison}
\begin{tabular}{lcccccc}
\toprule
\multirow{2}{*}{\textbf{Method}} 
& \multicolumn{3}{c}{\textit{Skywork-Reward-V2-Qwen3-0.6B}} 
& \multicolumn{3}{c}{\textit{Skywork-Reward-V2-Qwen3-1.7B}} \\ 
\cmidrule(lr){2-4} \cmidrule(lr){5-7}
 & \textbf{Top@1} & \textbf{Avg@N} & \textbf{Samples/s} 
 & \textbf{Top@1} & \textbf{Avg@N} & \textbf{Samples/s} \\ 
\midrule
BoN ($N=2$) & 1.51 & 0.77 & 0.59 & 2.09 & 1.18 & 0.59 \\
BoN ($N=4$) & 1.89 & 0.69 & 0.31 & 2.49 & 1.09 & 0.31 \\
APS ($M=1$) & 1.60 & 0.78 & 0.31 & 2.36 & 1.32 & 0.29 \\
APS ($M=3$) & 1.86 & 0.99 & 0.16 & 2.42 & 1.50 & 0.14 \\
\rowcolor{highlight} EntRGi ($M=1$) & 1.77 & 1.04 & 0.31 & 2.64 & 1.70 & 0.29 \\
\rowcolor{highlight} EntRGi ($M=3$) & 2.16 & 1.39 & 0.16 & 2.80 & 1.83 & 0.14 \\
\bottomrule
\end{tabular}
\end{table*}
\subsection{Throughput Analysis}
\label{subsec_appdx:throughput}

In~\autoref{tab:throughput_comparison}, we compare the throughput of standard gradient-free sampling (BoN) against gradient-guided approaches.  EntRGi incurs no overhead over APS, achieving throughput comparable to BoN ($N$=4). With similar throughput to BoN ($N$=4), EntRGi consistently outperforms it on the Avg@N metric across both reward models, and is competitive on the Top@1 metric. For gradient-based methods (APS, EntRGi), using a larger reward model leads to a slight reduction in throughput.

In~\autoref{fig:rl_experiment_appdx}, we visualize the speed/throughput of RGRL-EntRGi and RGRL-APS against diffu-GRPO. We observe an approximate 1.6$\times$ speedup with Dream on WildChat-IF. However, on the other three settings, RGRL-EntRGi and RGRL-APS are slower. This is expected as differentiation through the reward model trades off computational efficiency for sample efficiency. Our observations also suggest that the speedup may be setting dependent. In future work we aim to explore methods to more efficiently use the gradient feedback -- such as only during important decoding steps to better tradeoff sample-efficiency for compute efficiency. 

\begin{figure*}[h]
  \centering
  \includegraphics[width=\linewidth]{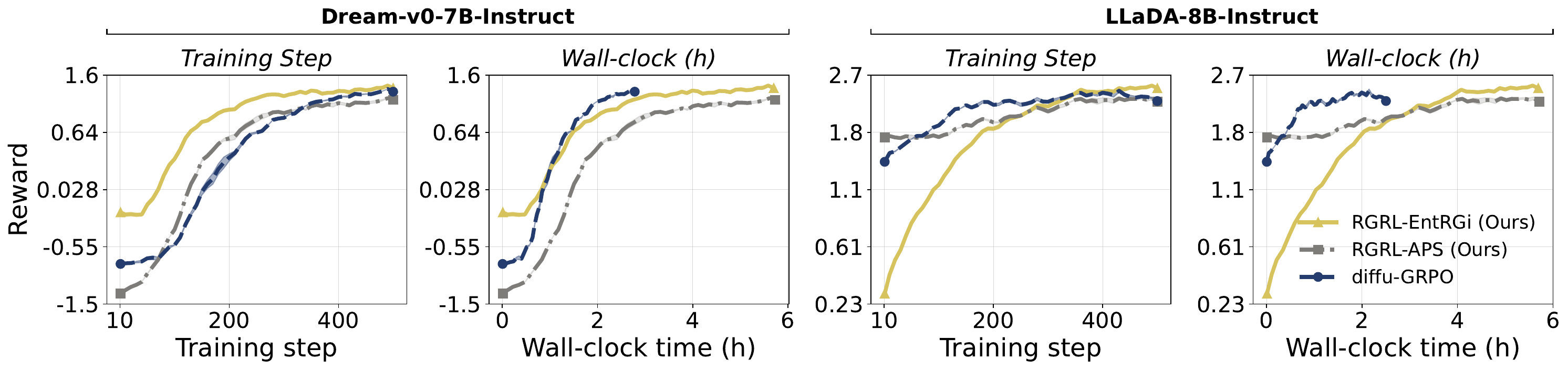}
  \caption{Training curves on Magpie-Ultra~\citep{xu2024magpiealignmentdatasynthesis}. Combined with \autoref{fig:rl_experiment}, we observe that RGRL's improvements scale inversely with the initial reward.}
  \label{fig:magpie_rl_experiment}
\end{figure*}

\begin{figure*}[h]
  \centering
  \includegraphics[width=\linewidth]{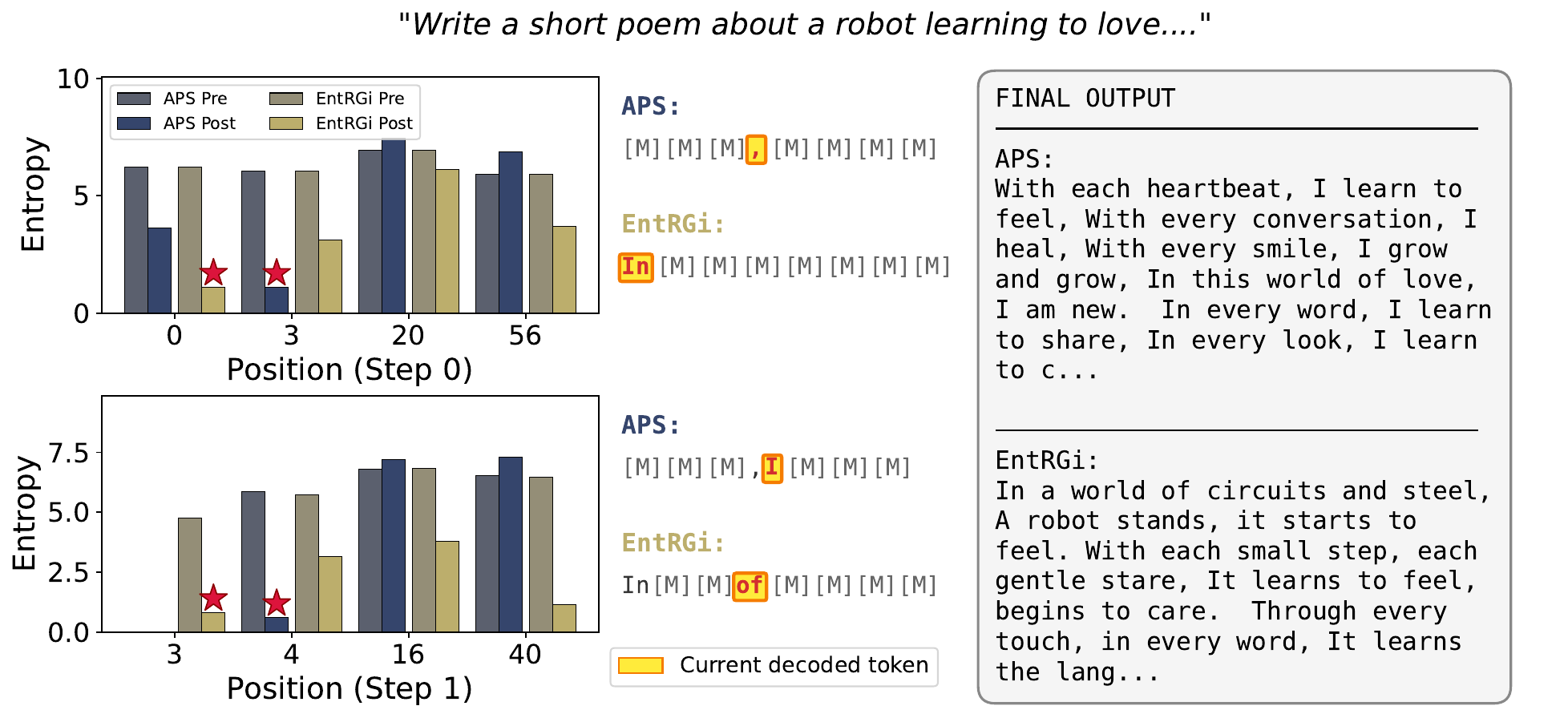}
  \caption{Qualitative example of APS vs. EntRGi.}
  \label{fig:qa1}
\end{figure*}
\begin{figure*}[h]
  \centering
  \includegraphics[width=\linewidth]{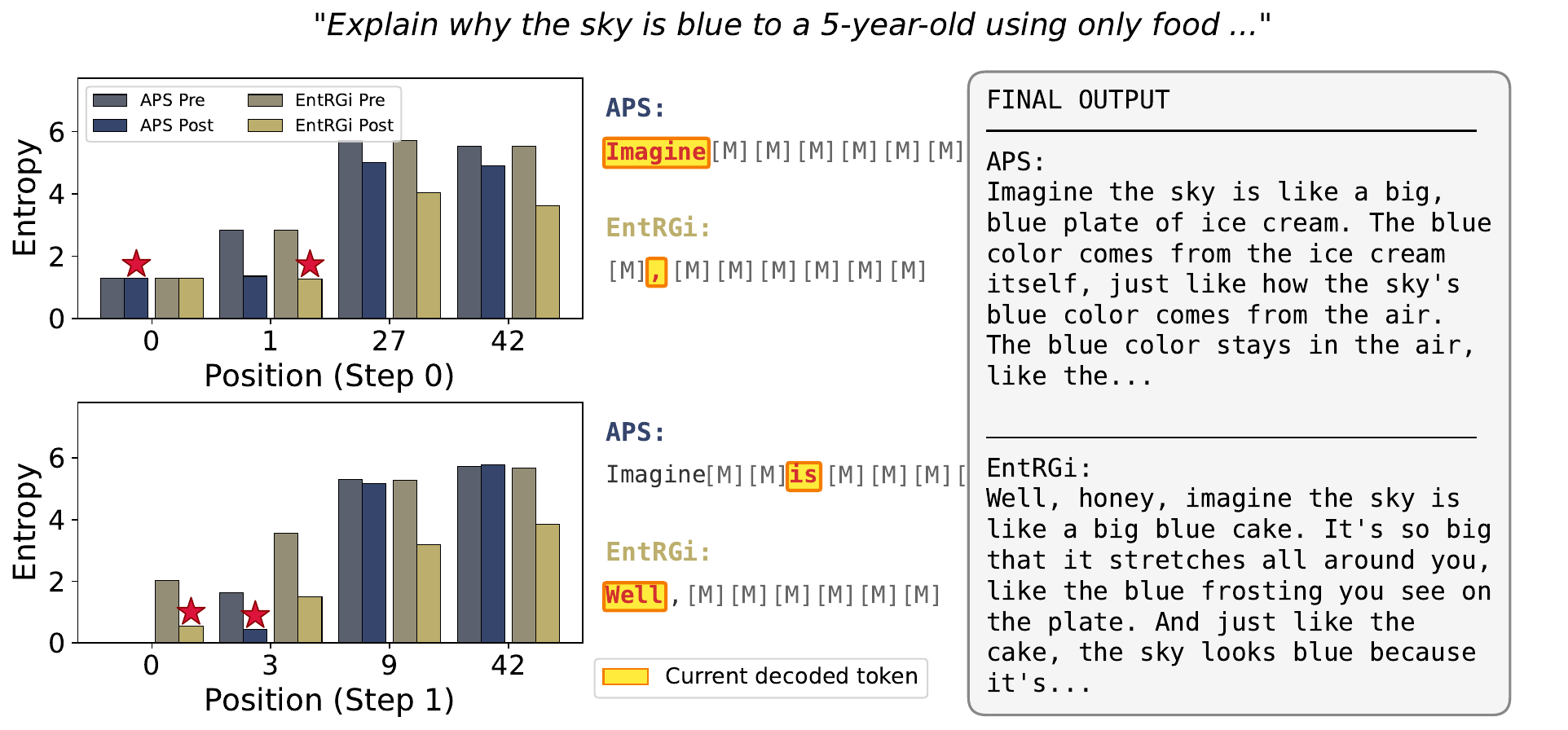}
  \caption{Qualitative example of APS vs. EntRGi}
  \label{fig:qa2}
\end{figure*}
\begin{figure*}[h]
  \centering
  \includegraphics[width=\linewidth]{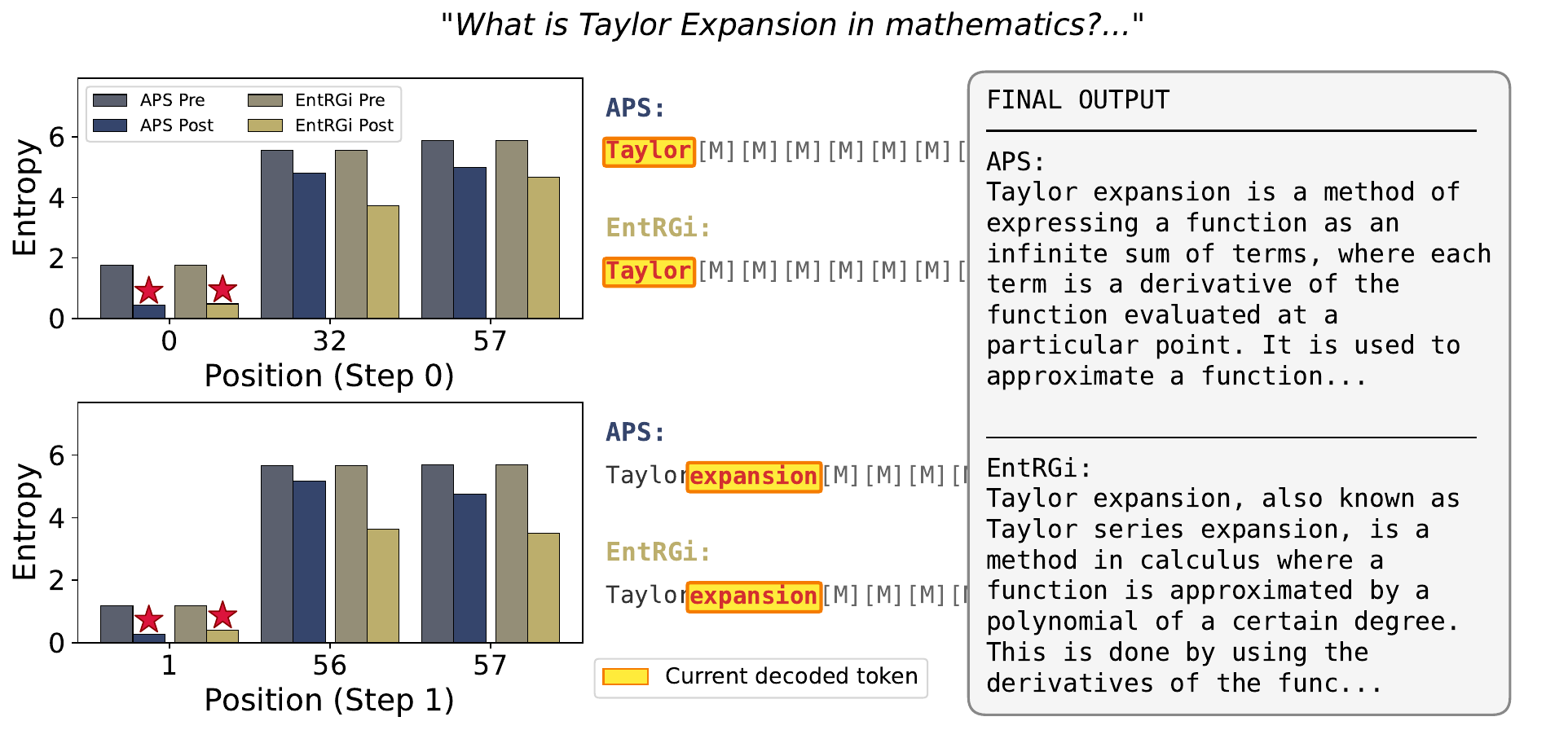}
  \caption{Qualitative example of APS vs. EntRGi}
  \label{fig:qa3}
\end{figure*}
\begin{figure*}[h]
  \centering
  \includegraphics[width=\linewidth]{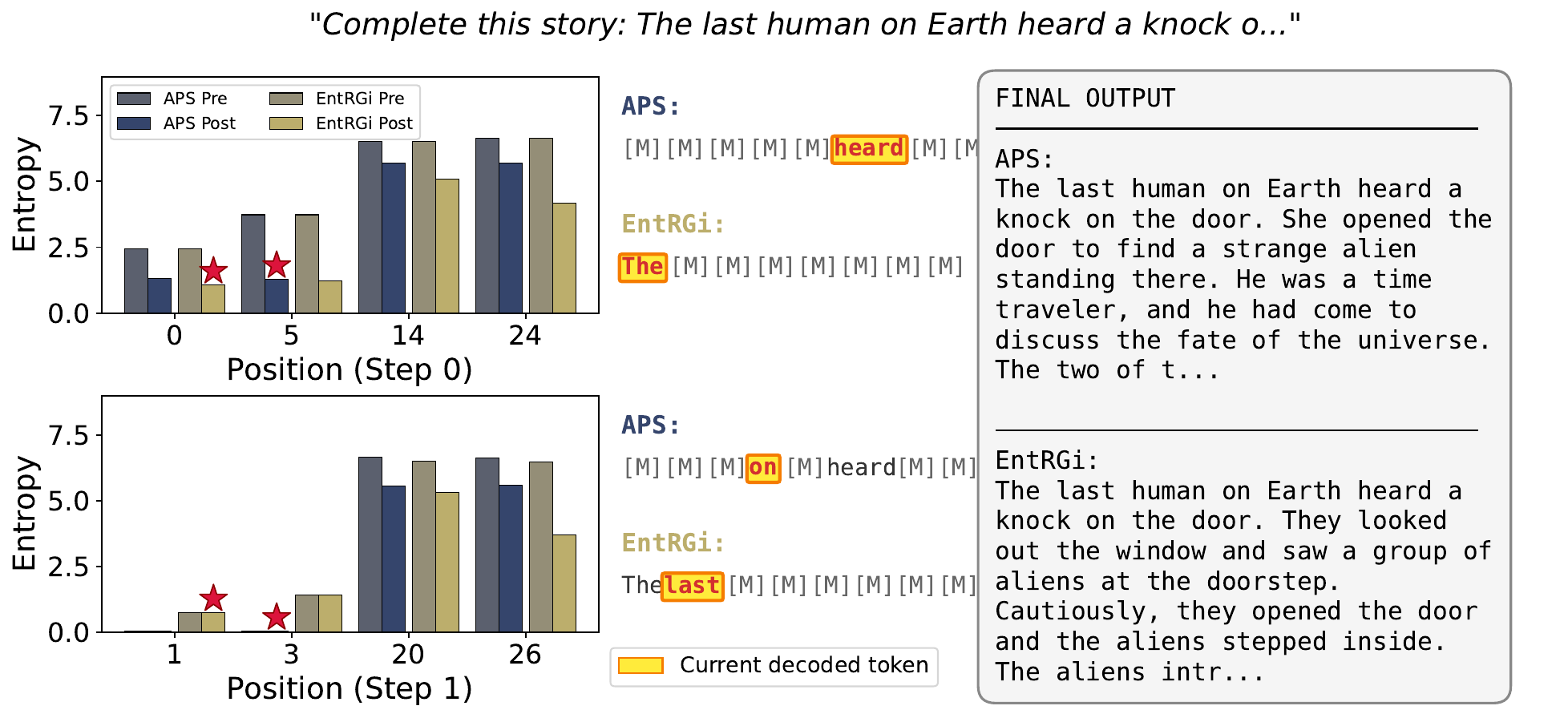}
  \caption{Qualitative example of APS vs. EntRGi}
  \label{fig:qa4}
\end{figure*}
\begin{figure*}[h]
  \centering
  \includegraphics[width=\linewidth]{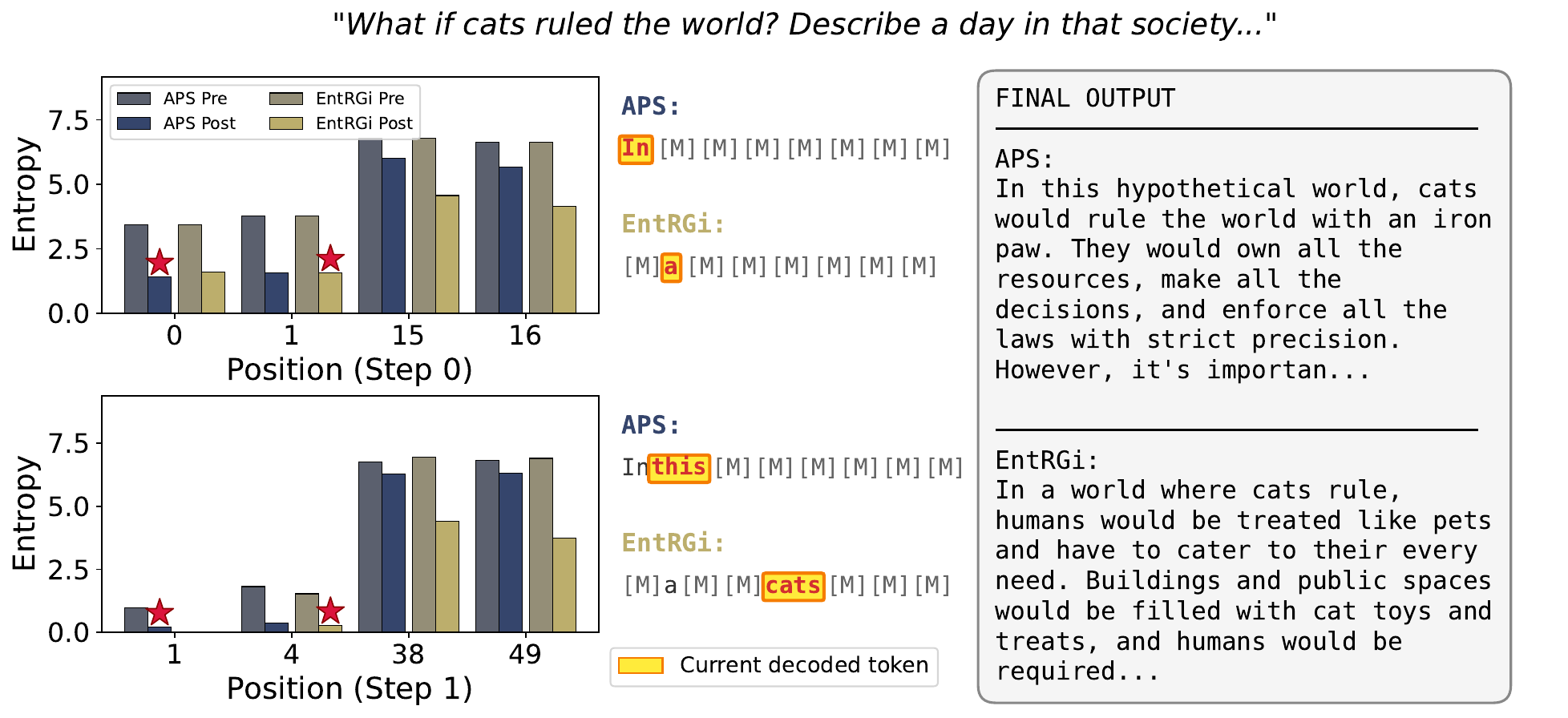}
  \caption{Qualitative example of APS vs. EntRGi}
  \label{fig:qa5}
\end{figure*}
\begin{figure*}[h]
  \centering
  \includegraphics[width=\linewidth]{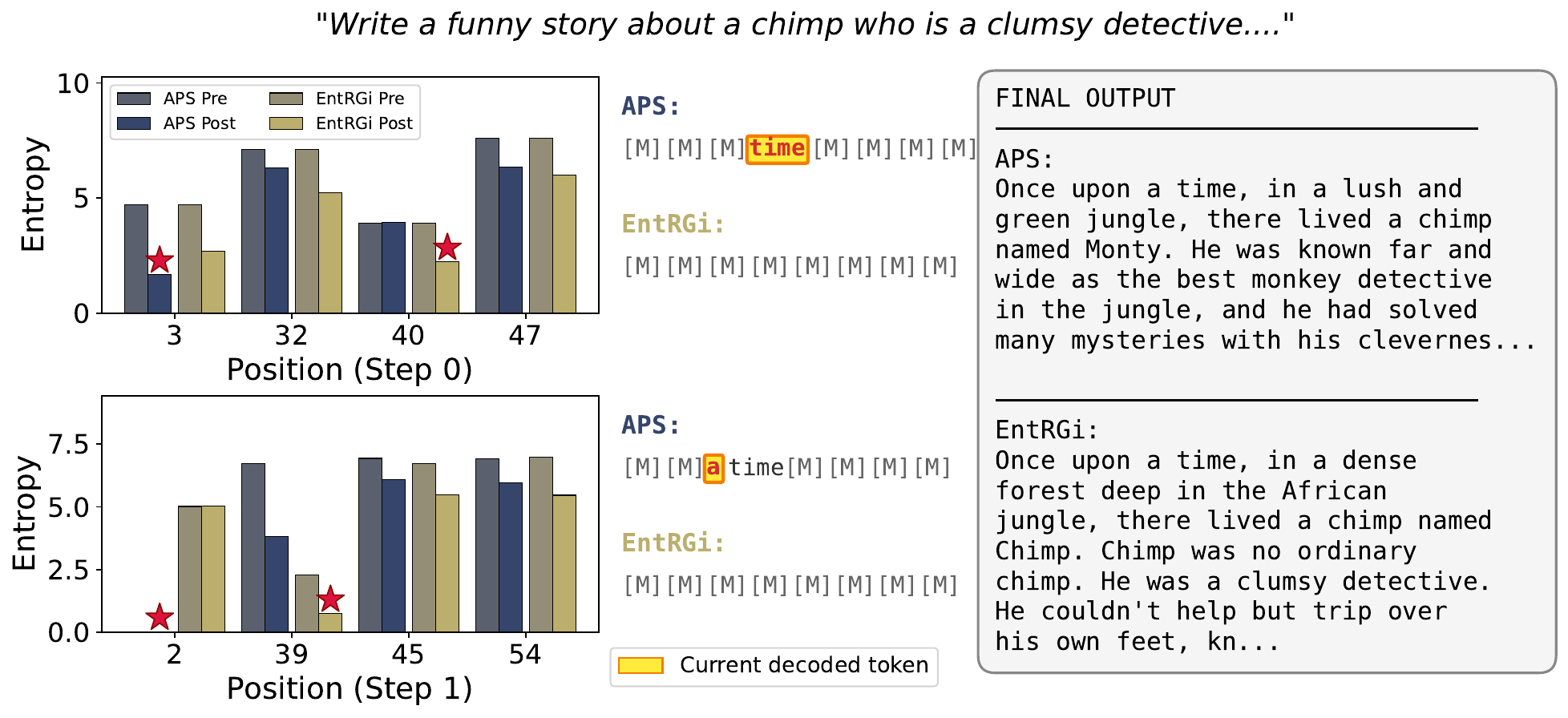}
  \caption{Qualitative example of APS vs. EntRGi}
  \label{fig:qa6}
\end{figure*}
\begin{figure*}[h]
  \centering
  \includegraphics[width=\linewidth]{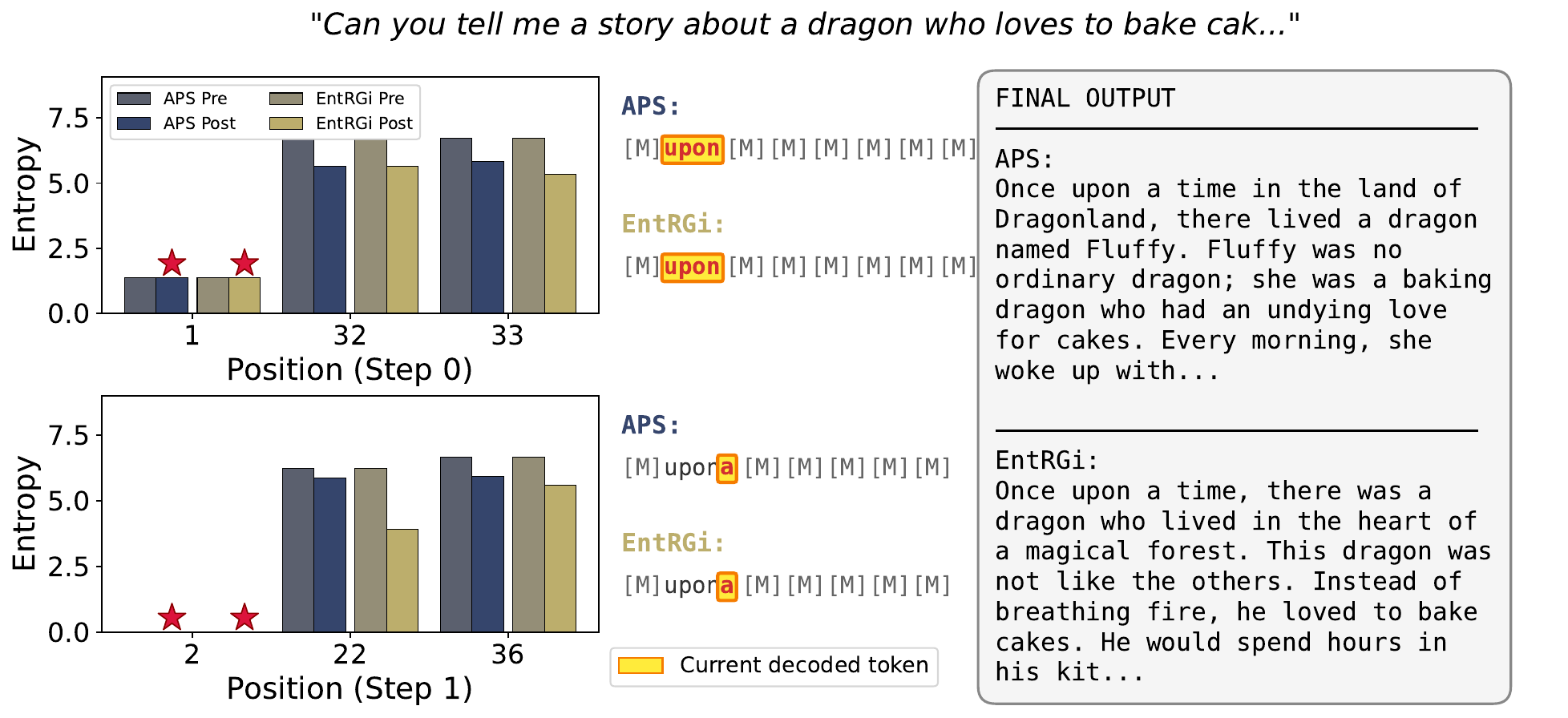}
  \caption{Qualitative example of APS vs. EntRGi}
  \label{fig:qa7}
\end{figure*}

\subsection{Qualitative Comparison}
\label{sec:appdx_qa}

We visualize and compare the generations of APS and EntRGi in \autoref{fig:qa1}, \autoref{fig:qa2}, \autoref{fig:qa3}, \autoref{fig:qa4}, \autoref{fig:qa5}, \autoref{fig:qa6}, and \autoref{fig:qa7}. All results are generated using a low temperature setting ($\tau = 0.1$) to minimize the effect of randomness in the final outputs. We observe several interesting behaviors across these examples.

Analyzing \autoref{fig:qa1}, the user asks for a short poem about a robot learning to love. The poem generated by APS is somewhat ambiguous, whereas EntRGi produces a more tailored poem that explicitly focuses on robotic themes.

In \autoref{fig:qa2}, the user asks for an explanation of the sky as if explaining it to a five-year-old. APS performs reasonably well by using analogies such as ice cream. EntRGi, however, captures finer-grained stylistic details, such as beginning with the phrase “Well, honey,” which adds a more personalized and engaging touch to the generation.

In \autoref{fig:qa5}, the user asks for a story about cats ruling the world. APS makes minimal use of cat-related analogies, while EntRGi includes richer thematic details, such as references to cat toys, treats, and humans catering to them.

Analyzing \autoref{fig:qa6}, the user requests a story about a chimp who is a clumsy detective. In the APS output, there is little indication of the chimp’s clumsiness, whereas EntRGi consistently incorporates this trait into the narrative.

\end{document}